\documentclass[11pt]{article}

\usepackage[preprint]{acl}

\usepackage{times}
\usepackage{latexsym}

\usepackage[T1]{fontenc}

\usepackage[utf8]{inputenc}

\usepackage{microtype}

\usepackage{inconsolata}

\usepackage{graphicx}

\usepackage{amsmath}
\usepackage{amssymb}
\usepackage{makecell}
\usepackage{booktabs}
\usepackage{listings}
\usepackage{xcolor} 
\usepackage{multirow}   
\usepackage{array}      

\newcommand{\vpara}[1]{\vspace{1.5ex}\noindent\textbf{#1}}

\title{AsyncTLS: Efficient Generative LLM Inference with Asynchronous Two-level Sparse Attention}

\author{
    Yuxuan Hu\textsuperscript{\rm 1,\rm 2},\ Jianchao Tan\textsuperscript{\rm 2},\ Jiaqi Zhang\textsuperscript{\rm 2},\ Wen Zan\textsuperscript{\rm 2}, Pingwei Sun\textsuperscript{\rm 2} \\
    \  {\bf Yifan Lu}\textsuperscript{\rm 2},
    \ {\bf Yerui Sun}\textsuperscript{\rm 2},\ {\bf Yuchen Xie}\textsuperscript{\rm 2},\ {\bf Xunliang Cai}\textsuperscript{\rm 2},\ {\bf Jing Zhang}\textsuperscript{\rm 1}\thanks{\ \ Corresponding author.} \\
    \textsuperscript{\rm 1}School of Information, Renmin University of China, Beijing, China \\
    \textsuperscript{\rm 2}Meituan, Beijing, China \\
    \texttt{\{huyuxuan1999,zhang-jing\}@ruc.edu.cn}
}

\begin{document}
\maketitle
\begin{abstract}
Long-context inference in LLMs faces the dual challenges of quadratic attention complexity and prohibitive KV cache memory. While token-level sparse attention offers superior accuracy, its indexing overhead is costly; block-level methods improve efficiency but sacrifice precision. We propose AsyncTLS, a hierarchical sparse attention system that combines coarse-grained block filtering with fine-grained token selection to balance accuracy and efficiency, coupled with an asynchronous offloading engine that overlaps KV cache transfers with computation via temporal locality exploitation. Evaluated on Qwen3 and GLM-4.7-Flash across GQA, and MLA architectures, AsyncTLS achieves accuracy comparable to full attention while delivering \(1.2\times\)-\(10.0\times\) operator speedups and \(1.3\times\)-\(4.7\times\) end-to-end throughput improvements on 48k–96k contexts.
\end{abstract}

\section{Introduction}

Large Language Models (LLMs)~\cite{deepseek-v3, kimik2, longcat-flash} have demonstrated remarkable capabilities across diverse natural language processing tasks, from conversational AI to complex reasoning and code generation. However, deploying these models at scale remains severely constrained by the self-attention mechanism's quadratic computational complexity and linear memory growth. This bottleneck becomes particularly acute during the decoding phase, where the Key-Value (KV) cache storage dominates memory consumption. As sequence lengths extend to hundreds of thousands of tokens, the KV cache footprint grows proportionally, frequently exceeding high-bandwidth GPU memory capacity and necessitating expensive offloading to slower memory tiers.

Sparse attention mechanisms have emerged as a practical solution to mitigate the quadratic computational and memory costs inherent in long-context modeling. Existing approaches can be systematically categorized along two principal dimensions: granularity of sparsity (token-level versus block-level) and selection strategy (static versus dynamic). Static and token-level methods, such as H2O~\cite{h2o-zhang2023}, StreamingLLM~\cite{stramingllm}, and SnapKV~\cite{snapkv-li2024}, employ fixed patterns to retain individual tokens. While these approaches enable fine-grained control over token participation and allow precise preservation of semantically salient information, they inherently fail to adapt to evolving attention patterns during generation, rendering them suboptimal when contextual relevance shifts dynamically. Conversely, dynamic and block-level methods, including Quest~\cite{quest-tang2024} and InfLLM~\cite{infllm-xiao2024}, utilize dynamic selection strategies that operate on contiguous token chunks to reduce indexing overhead and improve hardware efficiency. However, the coarse-grained nature of block-level aggregation inevitably compromises attention precision by incorporating irrelevant tokens within selected blocks while potentially discarding critical information in unselected regions, thereby introducing retrieval noise and degrading model fidelity.

Recent advances, including Double-Sparsity~\cite{double-sparsity-yang2024} and Deepseek Sparse Attention (DSA)~\cite{deepseekv3.2}, have demonstrated that token-level sparse attention achieves superior accuracy compared to block-level approaches under equivalent token budgets. By identifying and retaining individual important tokens rather than entire blocks, these methods more precisely capture long-range dependencies and critical contextual information scattered throughout sequences. However, this accuracy improvement entails significant overhead: the runtime indexing cost for token-level selection substantially exceeds that of block-level methods. Each query token requires computing importance scores across all candidate tokens and selecting top-k elements, operations that become performance bottlenecks when executed at every decoding step.

\begin{figure*}[t]  
    \centering
    \includegraphics[width=\textwidth]{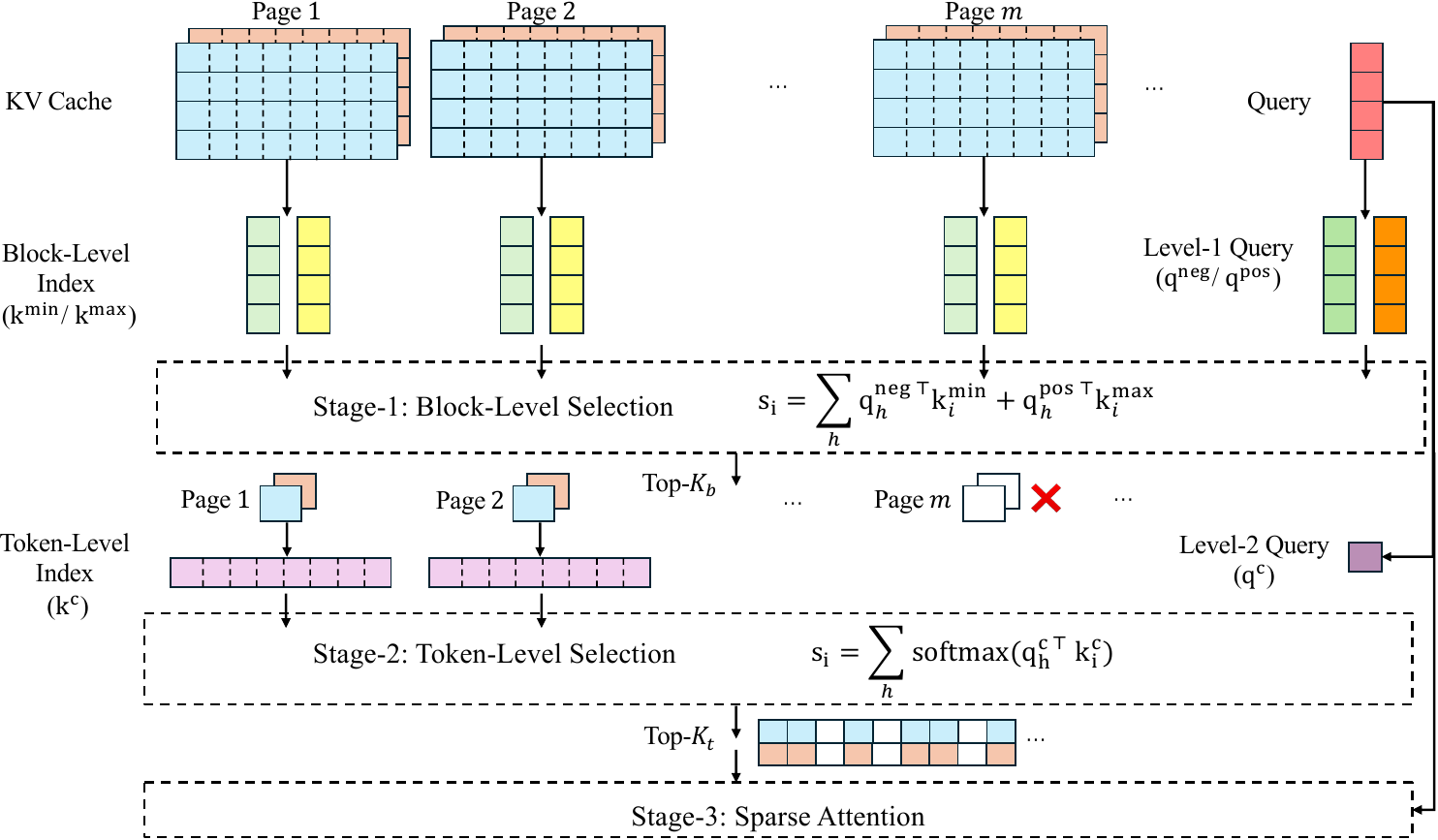}  
    \caption{Illustration of Two-Level Sparse Attention. We construct hierarchical indices for KV blocks at two granularities (Block-level and Token-level). The decoding process first selects Top-$K_b$ blocks via block-level indexing, then selects Top-$K_t$ tokens within these blocks via token-level indexing for sparse attention computation.}
    \label{fig: TLS}
\end{figure*}

This fundamental tension between accuracy and efficiency motivates our first contribution: a hierarchical two-level sparse attention architecture that synthesizes the strengths of both paradigms. Our approach employs block-level indexing as a coarse-grained filtering stage to rapidly eliminate sequence regions unlikely to contain relevant tokens, followed by token-level indexing to precisely select the most salient tokens within retained blocks for actual attention computation. This hierarchical design dramatically reduces the search space for fine-grained token selection while preserving the accuracy benefits of token-level sparsity.

Beyond the high computational overhead of attention mechanisms, the storage requirements of key-value caches may also exceed the limited capacity of high-bandwidth GPU memory. Consequently, KV cache offloading becomes essential when serving long-context workloads that risk surpassing GPU memory bounds. Prior work, including FlexGen~\cite{flexgen-sheng2023}, InfiniGen~\cite{infinigen-lee2024}, ShadowKV~\cite{shadowkv-sun2025}, and RetroInfer~\cite{retroinfer-chen2025}, has primarily focused on block-level sparsity patterns, transferring entire KV cache blocks between GPU and CPU memory. While effective for coarse-grained eviction, these approaches overlook optimization opportunities when combined with token-level sparse attention, particularly regarding fine-grained data movement and temporal locality exploitation.

Our second contribution addresses this gap by extending token-level sparse attention to the KV offloading scenario through an asynchronous prefetching mechanism. The key insight is that block-level filtering results from the current timestep serve as reliable predictors for token-level selection requirements in subsequent timesteps. Specifically, we employ a staggered execution strategy: at each decoding step, second-level token selection utilizes block filtering results from the previous step, while simultaneously prefetching KV blocks for the next step based on the current step's block filtering results. This design enables overlapping of KV transfer with attention computation, effectively hiding memory movement latency.

\begin{figure*}[t]  
    \centering
    \includegraphics[width=\textwidth]{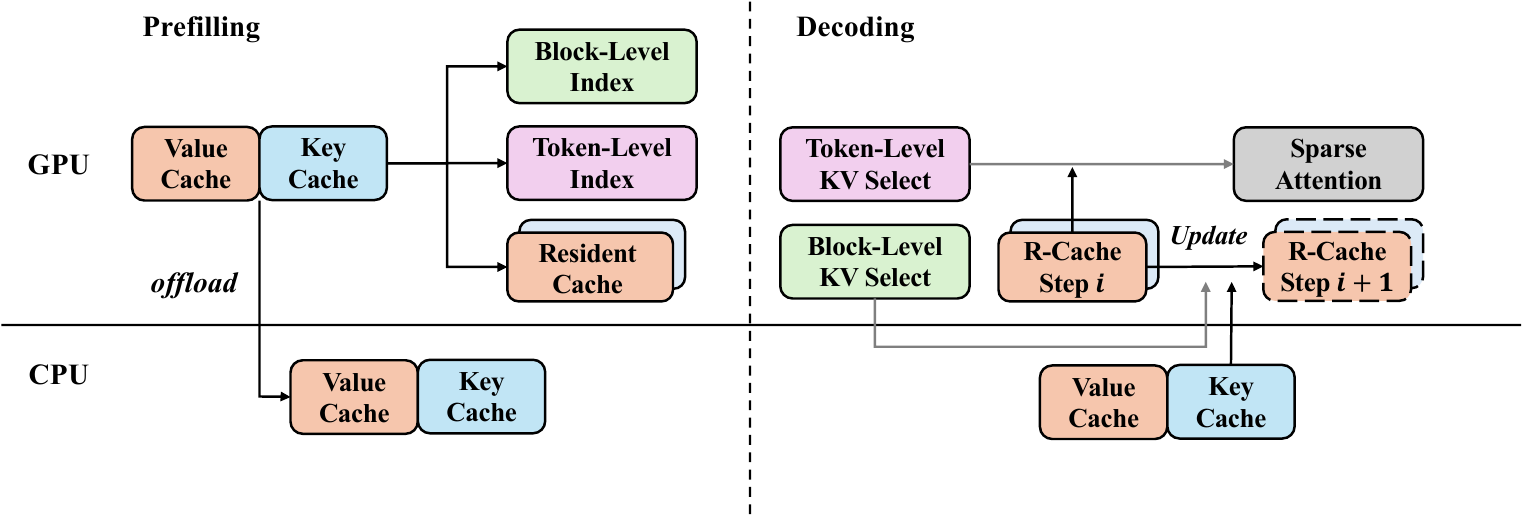}  
    \caption{Workflow of AsyncTLS. During prefilling, we first construct hierarchical indices for the input sequence. Based on the query representation of the final token, we retain critical KV pairs in the GPU resident cache and offload the remaining KV cache to the host (CPU) memory. During decoding, we perform sparse attention by retrieving KV pairs from the resident cache using token-level indexing. Concurrently, we asynchronously prefetch additional KV blocks from host memory via block-level indices to update the resident cache.}
    \label{fig:single}
\end{figure*}

Furthermore, we exploit temporal locality in token importance across adjacent timesteps. As critical token sets typically evolve gradually during decoding, we adopt an incremental KV block transfer strategy that transmits only blocks exhibiting selection divergence between consecutive steps. This approach eliminates redundant data movement while maximizing effective bandwidth utilization on the PCIe interconnect.

We evaluate AsyncTLS on state-of-the-art models, including Qwen3-8B, Qwen3-14B~\cite{qwen3-yang2025}, and GLM-4.7-Flash~\cite{glm4.5}, across comprehensive benchmarks spanning long-context retrieval and understanding tasks. Our evaluation spans Multi-Head Attention (MHA)~\cite{attention-vaswani2023} and Grouped-Query Attention (GQA)~\cite{gqa-shazeer2019} architectures, and we further extend validation to Multi-head Latent Attention (MLA)~\cite{deepseek-v2}, which employs compressed latent representations for KV caching, to demonstrate the broad architectural compatibility of our method. Under practical token budgets, AsyncTLS achieves accuracy nearly indistinguishable from full attention while outperforming existing baselines. Furthermore, end-to-end inference measurements demonstrate \(1.8\times\)–\(5.0\times\) operator speedups and \(1.3\times\)-\(1.8\times\) throughput improvements compared to Full Attention (FA) across context lengths ranging from 32k to 128k tokens.


In summary, this paper makes the following contributions:

\begin{itemize}
    \item Hierarchical Sparse Attention Architecture. We propose AsyncTLS, a two-level sparse attention mechanism combining coarse-grained block-level filtering with fine-grained token-level selection. This design achieves the accuracy benefits of token-level sparsity while mitigating prohibitive indexing overhead through hierarchical pruning.
    \item AsyncTLS Offloading Engine. We extend token-level sparse attention to the KV offloading setting through asynchronous prefetching and incremental block transfer, enabling efficient overlap of memory movement with computation while minimizing redundant data transmission across the memory hierarchy.
    \item Comprehensive Evaluation across Attention Paradigms. While existing sparse attention methods have primarily focused on MHA and GQA, our results demonstrate that training-free token-level sparsity maintains consistent effectiveness on MLA, achieving accuracy comparable to full attention under practical token budgets while delivering substantial inference speedups.

\end{itemize}

\section{Related Works}

In the domain of long-context inference optimization with static sparse pattern, H2O~\cite{h2o-zhang2023} introduces an eviction-based KV cache compression strategy via Heavy-Hitter Oracle. SnapKV~\cite{snapkv-li2024} further proposes an observation-window mechanism that identifies and compresses critical key-value pairs prior to generation. PyramidKV~\cite{pyramidkv-cai2025} introduces dynamic compression based on pyramidal information funneling, allocating larger cache budgets to lower layers where syntactic information concentrates. DynamicKV~\cite{dynamickv-zhou2025} observes that different tasks exhibit distinct activation patterns across layers, proposing a task-aware adaptive allocation. DuoAttention~\cite{duoattention-xiao} distinguishes between retrieval heads and streaming heads, applying full attention only to the former while maintaining rolling caches for the latter.

Regarding static pattern sparse attention mechanisms, QUEST~\cite{quest-tang2024} proposes query-aware dynamic sparsity that adaptively selects salient KV pairs based on attention distributions. \citep{double-sparsity-yang2024} present a post-training double-sparsity framework combining channel and token sparsity, enabling efficient inference without fine-tuning. Twilight~\cite{twilight-lin2025} designs hierarchical Top-p pruning with fine-grained gating to adjust per-head sparsity ratios dynamically. Star Attention~\cite{starattention-acharya2025} introduces a two-phase block-sparse mechanism that shards context processing across hosts with blockwise-local attention followed by sequence-global attention. For sampling and storage architectures, MagicPIG~\cite{magicpig-chen} employs Locality-Sensitive Hashing (LSH) sampling to approximate attention computation with linear complexity. FlexGen~\cite{flexgen-sheng2023}, InfiniGen~\cite{infinigen-lee2024}, ShadowKV~\cite{shadowkv-sun2025}, and RetroInfer~\cite{shadowkv-sun2025} combine sparse attention with a CPU-GPU hierarchical storage architecture that offloads secondary KV caches to host memory with asynchronous prefetching. 

Despite these advances, existing approaches incur prohibitive indexing costs for fine-grained selection and lack specialized offloading mechanisms for token-level sparsity, thereby failing to simultaneously achieve the accuracy of precise attention and the hardware efficiency required for practical ultra-long context deployment. To address these limitations, this paper presents AsyncTLS, a hierarchical sparse attention framework that bridges the granularity-efficiency gap through a two-level selection architecture, coupled with an asynchronous offloading engine optimized for dynamic token-level sparsity patterns.

\section{Preliminary}

\vpara{Multi-Head Attention and Its Variants.} The Transformer architecture relies on Multi-Head Attention (MHA), which computes scaled dot-product attention as:
\[
\text{Attention}(\mathbf{Q}, \mathbf{K}, \mathbf{V}) = \text{softmax}\left(\frac{\mathbf{Q}\mathbf{K}^\top}{\sqrt{d}}\right)\mathbf{V},
\]

where $n$ denotes the sequence length, $h$ the number of attention heads, and $d$ the dimension per head. Here, $\mathbf{Q}, \mathbf{K}, \mathbf{V} \in \mathbb{R}^{n \times d}$ are projected from the input $\mathbf{X} \in \mathbb{R}^{n \times d}$ via learned matrices $\mathbf{W}_Q, \mathbf{W}_K, \mathbf{W}_V \in \mathbb{R}^{d \times d}$. MHA employs $h$ parallel heads, where the $i$-th head computes $\mathbf{H}_i = \text{Attention}(\mathbf{X}\mathbf{W}_Q^i, \mathbf{X}\mathbf{W}_K^i, \mathbf{X}\mathbf{W}_V^i)$, and concatenates them as $\text{MHA}(\mathbf{X}) = \text{Concat}(\mathbf{H}_1, \dots, \mathbf{H}_h)\mathbf{W}_O$. During decoding, this requires caching $h$ key and value tensors, incurring $O(2hnd)$ memory.

To reduce this overhead, Multi-Query Attention (MQA) shares single key and value heads across all $h$ query heads, compressing the cache to $O(2nd)$. Grouped-Query Attention (GQA) generalizes this by grouping query heads to share KV heads, balancing efficiency and expressiveness. Multi-head Latent Attention (MLA) further compresses the cache via low-rank projections, and reconstructs keys/values during attention. During decoding, MLA effectively operates as a special case of MQA. By absorbing the projection matrices and utilizing the compressed latent representation, the multi key-value heads collapse into a single key-value head, minimizing memory while preserving representational capacity.


As sequences scale to hundreds of thousands of tokens, dense attention's quadratic complexity becomes prohibitive. Sparse attention reduces this complexity by restricting each query to attend only to a selected subset of keys and values. Formally, for the $i$-th query $\mathbf{q}_i \in \mathbb{R}^{d}$, an indexing function $\mathcal{I}(i) \subseteq \{1, \dots, n\}$ selects a subset of KV pairs, yielding $\mathbf{K}_{\mathcal{I}(i)}, \mathbf{V}_{\mathcal{I}(i)} \in \mathbb{R}^{|\mathcal{I}(i)| \times d}$. The sparse attention is then computed as:
\[
\text{SparseAttn}(\mathbf{q}_i, \mathbf{K}, \mathbf{V}) = \text{softmax}\left(\frac{\mathbf{q}_i\mathbf{K}_{\mathcal{I}(i)}^\top}{\sqrt{d}}\right)\mathbf{V}_{\mathcal{I}(i)},
\]

where the softmax is applied over the selected indices $\mathcal{I}(i)$, reducing computational complexity from $O(n^2d)$ to $O(n \cdot |\mathcal{I}(i)| \cdot d)$. The indexing strategy $\mathcal{I}(\cdot)$ can be static or dynamic, trading off between hardware efficiency and model fidelity.


\section{Method}
\label{sec:method}

We now present our method \textbf{AsyncTLS}, an efficient sparse attention mechanism designed for long-context inference with KV cache offloading. The core insight of AsyncTLS is twofold: (1) \textit{hierarchical sparsity} that combines the efficiency of block-level indexing with the precision of token-level selection, and (2) \textit{asynchronous prefetching} that overlaps KV cache transmission with attention and feed-forward network computation by exploiting temporal locality across decoding steps.

Given an input sequence of length $n$, we process the KV cache $\mathbf{K}, \mathbf{V} \in \mathbb{R}^{n \times d}$ through a two-level selection pipeline. At each decoding timestep $t$, we first identify relevant blocks using coarse-grained scoring, then apply fine-grained token pruning within selected blocks. To hide memory transfer latency during offloading, we asynchronously prefetch KV blocks based on predictions from previous timesteps and transmit only incremental differences between consecutive selections.

\subsection{Two-Level Sparse Attention}
\label{sec:two_level}

\paragraph{Coarse-grained Block Selection.}
Following Quest, we partition the KV cache into $m$ blocks of size $B$, where $m = \lfloor n/B \rfloor$. For query head $\mathbf{q}_h \in \mathbb{R}^{d}$, Quest computes block importance scores using each block's compressed representation:
\begin{align*}
    & \mathbf{\hat{k}}^{\max}_{i,k} = \max_{j\in \mathcal{B}_i} (k_{j,k}), \\
    & \mathbf{\hat{k}}^{\min}_{i,k} = \min_{j\in \mathcal{B}_i} (k_{j,k}), \\
    & s_i = \sum_{h=1}^G \sum_{k =1}^{d} \max(\mathbf{q}_{h,k} \mathbf{k}^{\max}_{i,k}, \mathbf{q}_{h,k} \mathbf{k}^{\min}_{i,k}),
\end{align*}

\noindent where $G$ the number of query heads sharing the same key, $\mathcal{B}_i$ denotes the $i$-th block and $\mathbf{\hat{k}}^{\max}_i$, $\mathbf{\hat{k}}^{\min}_i$ represents compressed block representations. 

While Quest achieves effective block-level selection, its importance scoring computation resists efficient mapping to dense matrix multiplication (GEMM) primitives. Although this inefficiency incurs only modest overhead on architectures with independent KV projections (e.g., MHA and GQA), it severely undermines compute unit utilization, particularly Tensor Cores, for shared-KV architectures such as MQA and MLA. To address this architectural mismatch, we reformulate Quest's scoring mechanism into standard GEMM operations, thereby fully exploiting the computational capabilities of modern accelerators. Concretely, for block $i$, the importance score can also be computed as:
\begin{align*}
    & \mathbf{\hat{k}}^{\max}_{i,k} = \max_{j\in \mathcal{B}_i} (k_{j,k}), \\
    & \mathbf{\hat{k}}^{\min}_{i,k} = \min_{j\in \mathcal{B}_i} (k_{j,k}), \\
    & \mathbf{q}^{\max}_{k} = \max(\mathbf{q}_{k}, 0), \\
    & \mathbf{q}^{\min}_{k} = \min(\mathbf{q}_{k}, 0), \\
    & s_i = \sum_{h=1}^G \left( \mathbf{q}^{\max\top}_{h} \mathbf{\hat{k}}^{\max}_{i} + \mathbf{q}^{\min\top}_{h} \mathbf{\hat{k}}^{\min}_{i} \right),
\end{align*}

\noindent which can be expressed compactly for all blocks via matrix multiplication:
\begin{equation*}
    \mathbf{s} = \sum_{h=1}^G\left( \mathbf{Q}^{\max\top}\mathbf{K}^{\max} + \mathbf{Q}^{\min\top}\mathbf{K}^{\min} \right)_h,
\end{equation*}

\noindent where $\mathbf{Q}^{\max/\min} \in \mathbb{R}^{d_k \times G}$ and $\mathbf{K}^{\max/\min} \in \mathbb{R}^{d_k \times N_b}$ denote the aggregated max/min query and key representations across all heads and blocks, respectively. We select the top-$k_b$ blocks $\mathcal{M}_t$ with the highest scores at timestamp $t$ for each key-value group, forming a coarse candidate set that retains $k_b \cdot B$ tokens.

\paragraph{Fine-grained Token Selection.}
Within the selected blocks $\mathcal{M}_t$, we apply Double Sparsity to perform token-level selection. To identify the most informative channels for attention score approximation, we first perform calibration on a held-out dataset $\mathcal{D}_{\text{cal}}$. Let $G$ denote the number of query heads sharing the same key for each channel $i \in \{1, \dots, d\}$, we compute its importance score by aggregating the maximum absolute values across all heads:
\begin{equation*}
    s_i = \frac{1}{G} \sum_{h=1}^{G} \left(\max_{(\mathbf{q}, \mathbf{k}) \in \mathcal{D}_{\text{cal}}} |\mathbf{q}_h[i]|\right) \cdot \left(\max_{(\mathbf{q}, \mathbf{k}) \in \mathcal{D}_{\text{cal}}} |\mathbf{k}[i]|\right),
\end{equation*}
where $\mathbf{q}_h$ and $\mathbf{k}$ denote the query and the corresponding key for the $h$-th head. We then select the top-$d_c$ channels with the highest scores to form the representative channel set $\mathcal{C}$.

We compress the query and key vectors by projecting them onto these selected channels. Additionally, considering the substantial GPU memory overhead associated with token-level indexing, we further combined quantization with channel selection to compress the key vectors:
\begin{equation*}
    \tilde{\mathbf{q}}^{(h)} = \mathbf{q}^{(h)}[:, \mathcal{C}], \quad \tilde{\mathbf{k}}_j = \text{Quantize}(\mathbf{k}_j[:, \mathcal{C}]),
\end{equation*}
where $\tilde{\mathbf{q}}_t^{(h)}$ denotes the compressed query vector of the $h$-th head within the group. We approximate the attention score by averaging across the $G$ query heads sharing the same key:
\begin{equation*}
    \tilde{\alpha}_j = \frac{1}{G} \sum_{h=1}^{G} \text{softmax}( \frac{\tilde{\mathbf{q}}^{(h)} \tilde{\mathbf{k}}_j^\top}{\sqrt{d}}).
\end{equation*}
We then select the top-$k_t$ tokens based on these approximate scores:
\begin{equation*}
    \mathcal{S}_t = \text{TopK}_{k_t}\left(\{\tilde{\alpha}_j \mid j \in \mathcal{M}_t\}\right).
\end{equation*}

\noindent Finally, the attention output for each head $h$ is computed using the full-dimensional KV pairs indexed by $\mathcal{S}_t$:
\begin{equation*}
    \mathbf{o}_t^{(h)} = \text{softmax}\left(\frac{\mathbf{q}^{(h)} \mathbf{K}_{\mathcal{S}_t}^\top}{\sqrt{d}}\right) \mathbf{V}_{\mathcal{S}_t},
\end{equation*}
where $\mathbf{K}_{\mathcal{S}_t}, \mathbf{V}_{\mathcal{S}_t}$ denote the full-dimensional KV pairs indexed by $\mathcal{S}_t$.

\begin{table*}[t]
\newcolumntype{?}{!{\vrule width 1pt}}
\newcolumntype{C}{>{\centering\arraybackslash}p{1.2cm}}
\centering
\renewcommand{\arraystretch}{1.2}
\resizebox{\textwidth}{!}{
\begin{tabular}{c?c?ccccccccccccc?c}
\toprule
                     &          & \multicolumn{3}{c?}{Single-Doc QA}       & \multicolumn{3}{c?}{Multi-Doc QA}        & \multicolumn{3}{c?}{Summarization}       & \multicolumn{1}{c?}{Few-shot} & \multicolumn{1}{c?}{Synthetic} & \multicolumn{2}{c?}{Code} & Avg. \\ 
                     &          & NQA   & QQA   & \multicolumn{1}{c?}{MFQ} & HQA   & 2WM   & \multicolumn{1}{c?}{Mus} & GvR   & QMS   & \multicolumn{1}{c?}{MNs} & \multicolumn{1}{c?}{TQA}      & \multicolumn{1}{c?}{PRetr}     & LCC         & RBP         &      \\ \midrule
\multirow{4}{*}{\makecell{Qwen3-8B \\ (32k)}}
                     & Full     & 25.61 & 44.17 & 53.40                    & 53.48 & 38.29 & 32.14                    & 33.17 & 23.53 & 24.93                    & 90.71                         & 100.0                          & 67.56       & 64.92       &  50.14    \\
                     \cline{3-16}
                     & Quest    & 20.64 & 40.13 & 51.01                    & 45.74 & \textbf{38.46} & 27.25                    & 32.13 & 22.21 & \textbf{24.95}                    & 87.55                         & 98.5                           & 66.86       & 61.95       &  47.49    \\
                     & DS       & 22.50 & \textbf{44.76} & \textbf{54.05}                    & \textbf{54.22} & 36.26 & \textbf{33.13}                    & 33.51 & 23.47 & 24.65                    & \textbf{90.06}                         & \textbf{100.0}                          & \textbf{68.85}       & 65.94       &  \textbf{50.10}    \\
                     & AsyncTLS & \textbf{24.77} & 44.35 & 52.67                    & 53.59 & 38.02 & 31.1                     & \textbf{33.58} & \textbf{23.67} & 24.72                    & 89.88                         & 99.5                           & 67.67       & \textbf{65.17}       &  49.90    \\ \midrule
\multirow{4}{*}{\makecell{Qwen3-14B \\ (32k)}}
                     & Full     & 27.77 & 44.58 & 49.53                    & 60.93 & 48.9  & 36.02                    & 33.29 & 23.59 & 24.89                    & 92.25                         & 100.0                          & 69.75       & 66.63       & 52.16     \\
                     \cline{3-16}
                     & Quest    & 23.15 & 43.01 & 47.88                    & 54.93 & 44.56 & 32.43                    & 32.03 & 22.68 & 24.83                    & 90.41                         & 99.5                           & 67.53       & 58.31       & 49.33     \\
                     & DS       & \textbf{26.44} & \textbf{44.03} & \textbf{49.93}                    & \textbf{61.35} & 45.52 & 27.25                    & 33.83 & 24.06 & \textbf{24.93}                    & \textbf{92.75}                         & \textbf{100.0}                          & 69.83       & 64.40       & 51.10     \\
                     & AyncTLS  & 25.07 & 43.95 & 49.87                    & 60.97 & \textbf{47.86} & \textbf{35.16}                    & \textbf{33.34} & \textbf{24.16} & 24.87                    & 92.25                         & \textbf{100.0}                          & \textbf{70.26}       & \textbf{65.65}       & \textbf{51.80}     \\ \midrule
\multirow{4}{*}{\makecell{GLM-4.7-Flash \\ (128k)}}
                     & Full     & 26.8  & 36.17 & 56.74                    & 57.86 & 43.88 & 31.04                    & 32.87 & 23.4  & 26.87                    & 92.07                         & 99.0                           & 69.79       & 64.28       & 50.83    \\
                     \cline{3-16}
                     & Quest    & 25.01 & 33.77 & 52.21                    & 44.99 & 36.83 & 23.03                    & 31.29 & 21.87 & 26.29                    & 89.29                         & 96.0                           & 65.4        & 61.77       & 46.75    \\
                     & DS       & 26.68 & 35.53 & 55.68                    & \textbf{60.03} & \textbf{45.91} & \textbf{30.66}                    & \textbf{32.77} & \textbf{23.04} & 26.56                    & 91.09                         & 98.5                           & \textbf{68.33}       & 64.75       & 50.73    \\
                     & AsyncTLS & \textbf{27.24} & \textbf{36.71} & \textbf{56.15}                    & 56.03 & 44.28 & 30.37                    & 32.55 & 22.96 & \textbf{26.67}                    & \textbf{91.84}                         & \textbf{99.0}                           & 68.22       & \textbf{65.55}       & \textbf{50.58}    \\
\bottomrule
\end{tabular}
}
\caption{Experiment results of AsyncTLS and baseline methods on LongBench.\label{exp: longbench}}
\end{table*}

\begin{table*}[t]
\newcolumntype{?}{!{\vrule width 1pt}}
\newcolumntype{C}{>{\centering\arraybackslash}p{1.2cm}}
\centering
\renewcommand{\arraystretch}{1.2}
\resizebox{0.85\textwidth}{!}{
\begin{tabular}{c?c?cccccccccc?c}
\toprule
          &          & S1    & S2    & MK1   & MK2   & MQ    & MV    & QA-1  & QA-2  & VT    & FWE   & Avg. \\ \midrule
\multirow{4}{*}{\makecell{Qwen3-8B}}
          & Full     & 100.0 & 100.0 & 99.60 & 98.20 & 99.75 & 99.60 & 49.00 & 60.90 & 99.80 & 97.67 & 90.45     \\
          \cline{3-13}
          & Quest    & \textbf{100.0} & 30.20 & 28.00 & 4.00  & 11.25 & 5.40  & 32.60 & 39.35 & 50.48 & \textbf{87.87} & 38.92     \\
          & DS       & \textbf{100.0} & \textbf{100.0} & 99.60 & 98.20 & 99.65 & \textbf{99.75} & 49.20 & \textbf{62.92} & 99.76 & 95.27 & 90.44     \\
          & AsyncTLS & \textbf{100.0} & \textbf{100.0} & \textbf{99.80} & \textbf{98.60} & \textbf{99.70} & 99.45 & \textbf{52.20} & 59.62 & \textbf{99.80} & 96.27 & \textbf{90.54}     \\ \midrule
\multirow{4}{*}{\makecell{Qwen3-14B}}
          & Full     & 100.0 & 100.0 & 99.80 & 99.60 & 99.95 & 100.0 & 57.80 & 66.60 & 100.0 & 38.67 &  86.24    \\
          \cline{3-13}
          & Quest    & \textbf{100.0} & 34.60 & 41.00 & 5.00  & 17.85 & 10.20 & 35.60 & 44.37 & 40.24 & 50.00 &  37.89    \\
          & DS       & \textbf{100.0} & 34.60 & \textbf{100.0} & \textbf{99.80} & \textbf{99.95} & \textbf{99.85} & 55.60 & \textbf{65.70} & 99.92 & 16.60 &  77.20    \\
          & AsyncTLS & \textbf{100.0} & \textbf{100.0} & 99.80 & 99.60 & 99.80 & 99.65 & \textbf{58.40} & 64.75 & \textbf{100.0} & \textbf{50.13} &  \textbf{87.21}   \\ \midrule
\end{tabular}
}
\caption{Experiment results of AsyncTLS and baseline methods on RULER for Qwen3 models.\label{exp: ruler qwen}}

\end{table*}


\begin{table}[t]
\newcolumntype{?}{!{\vrule width 1pt}}
\newcolumntype{C}{>{\centering\arraybackslash}p{1.2cm}}
\centering
\renewcommand{\arraystretch}{1.2}
\resizebox{\linewidth}{!}{
\begin{tabular}{c?cccccc?c}
\toprule
                   & MQ    & MV    & QA-1  & QA-2  & VT    & FWE   & Avg. \\ \midrule
          Full     & 99.55 & 99.75 & 53.60 & 66.25 & 100.0 &     95.20 & 85.73     \\
          \cline{2-8}
          Quest    & 57.45 & 42.25 & 30.40 & 45.80 & 60.84 &     83.20 & 53.32     \\
          DS       & 92.90 & \textbf{99.75} & \textbf{49.00} & 59.40 & \textbf{100.0} &     \textbf{93.20} & \textbf{82.38}     \\
          AsyncTLS & \textbf{95.10} & 98.80 & 48.20 & \textbf{59.50} & 99.88 &     87.53  & 81.50    \\
\bottomrule
\end{tabular}
}
\caption{Experiment results of AsyncTLS and baseline methods on RULER for GLM-4.7-Flash.\label{exp: ruler glm}}
\end{table}

\begin{figure*}[t]
    \centering
    \begin{minipage}[b]{0.23\textwidth}
        \centering
        \includegraphics[width=\textwidth]{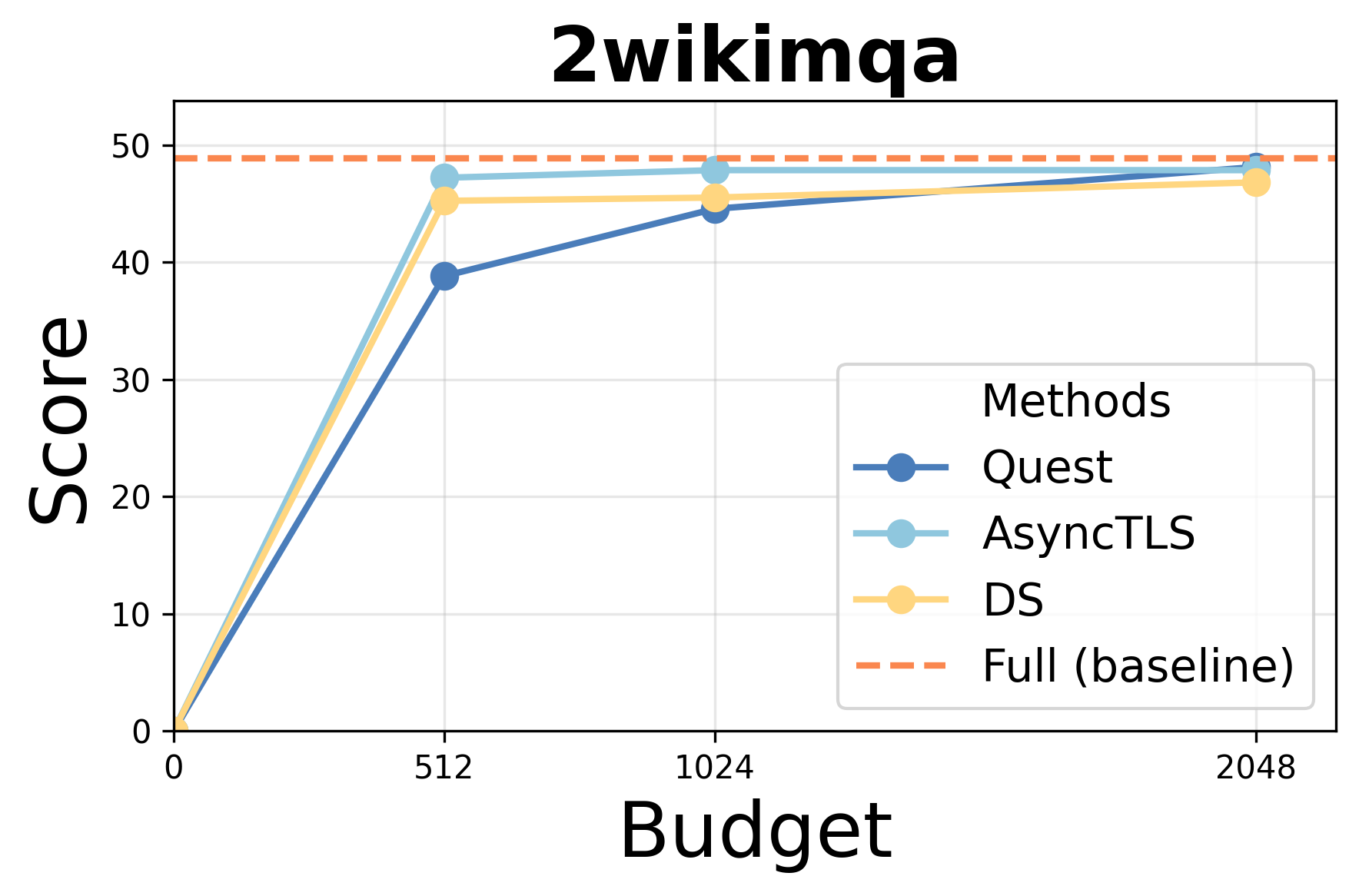}
    \end{minipage}
    \begin{minipage}[b]{0.23\textwidth}
        \centering
        \includegraphics[width=\textwidth]{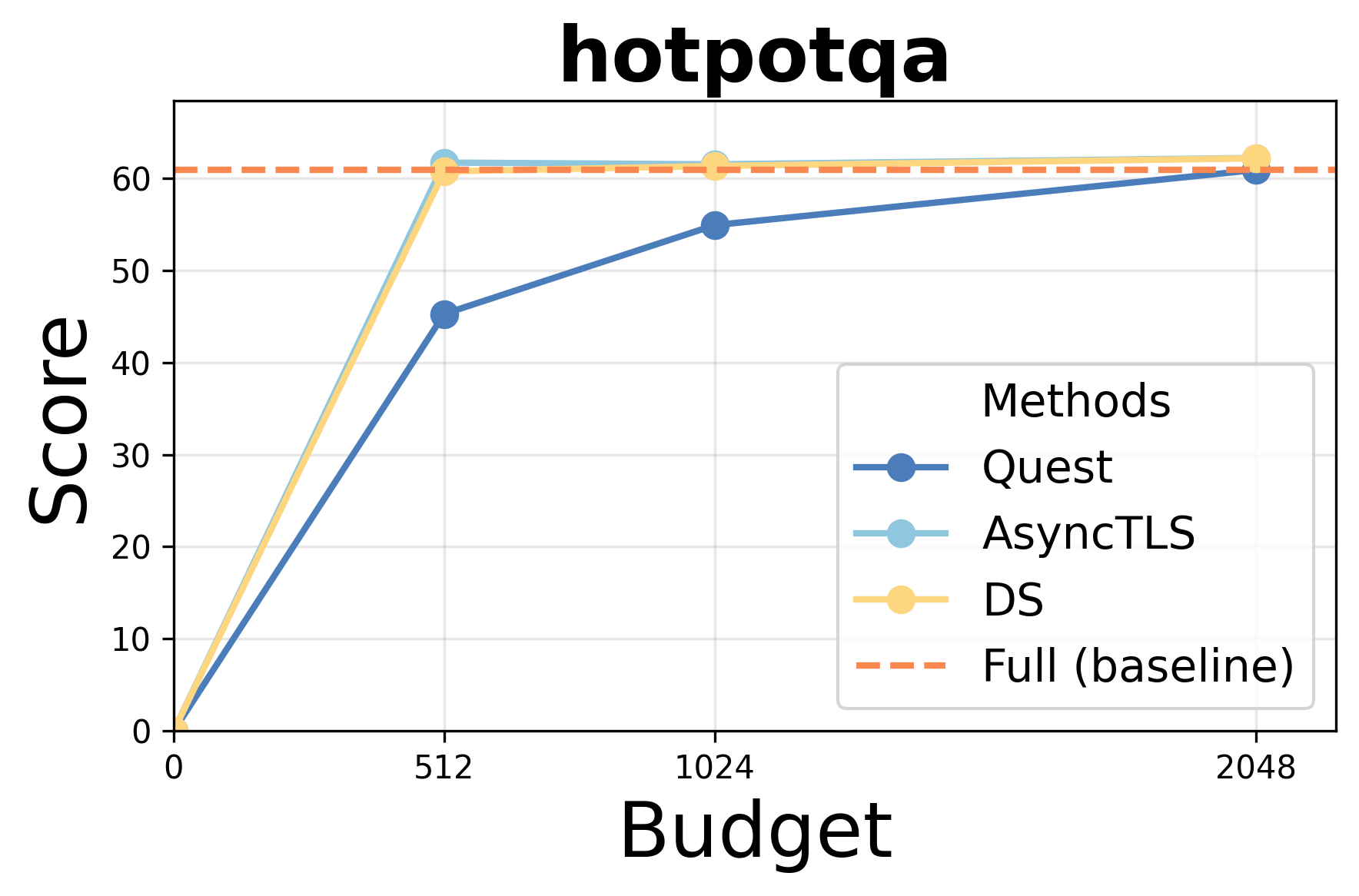}
    \end{minipage}
    \begin{minipage}[b]{0.23\textwidth}
        \centering
        \includegraphics[width=\textwidth]{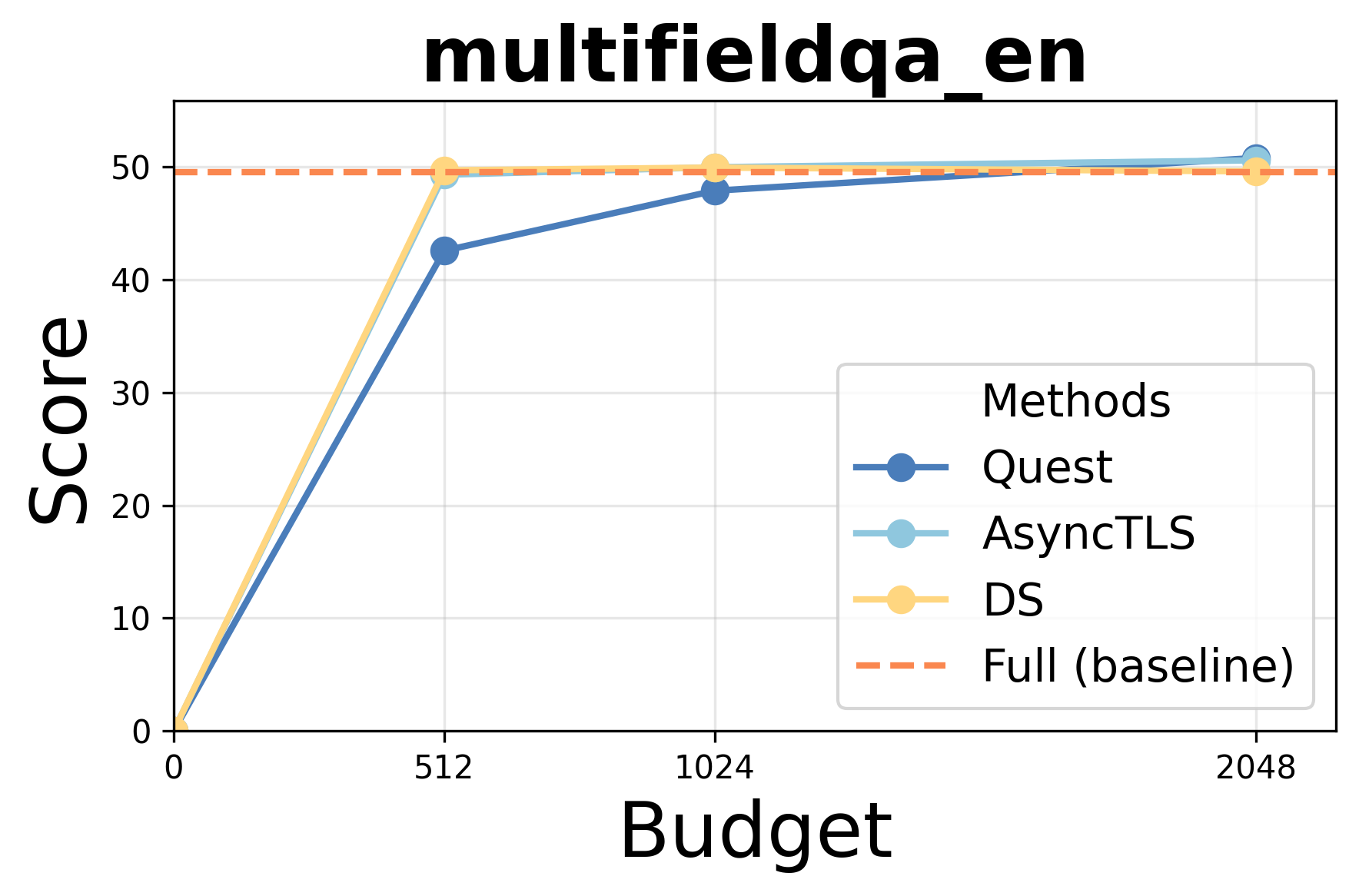}
    \end{minipage}
    \begin{minipage}[b]{0.23\textwidth}
        \centering
        \includegraphics[width=\textwidth]{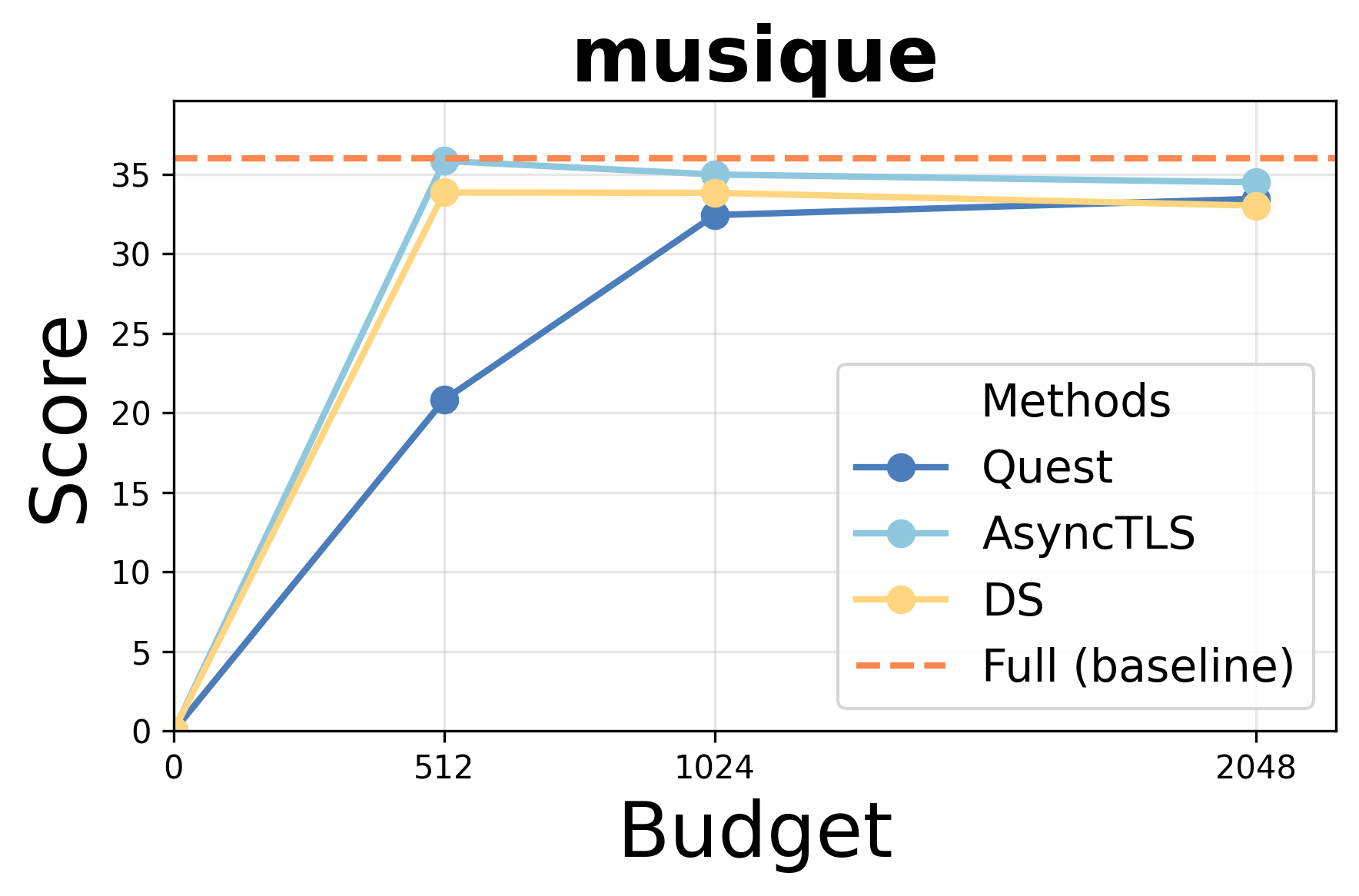}
    \end{minipage}
        
    \begin{minipage}[b]{0.23\textwidth}
        \centering
        \includegraphics[width=\textwidth]{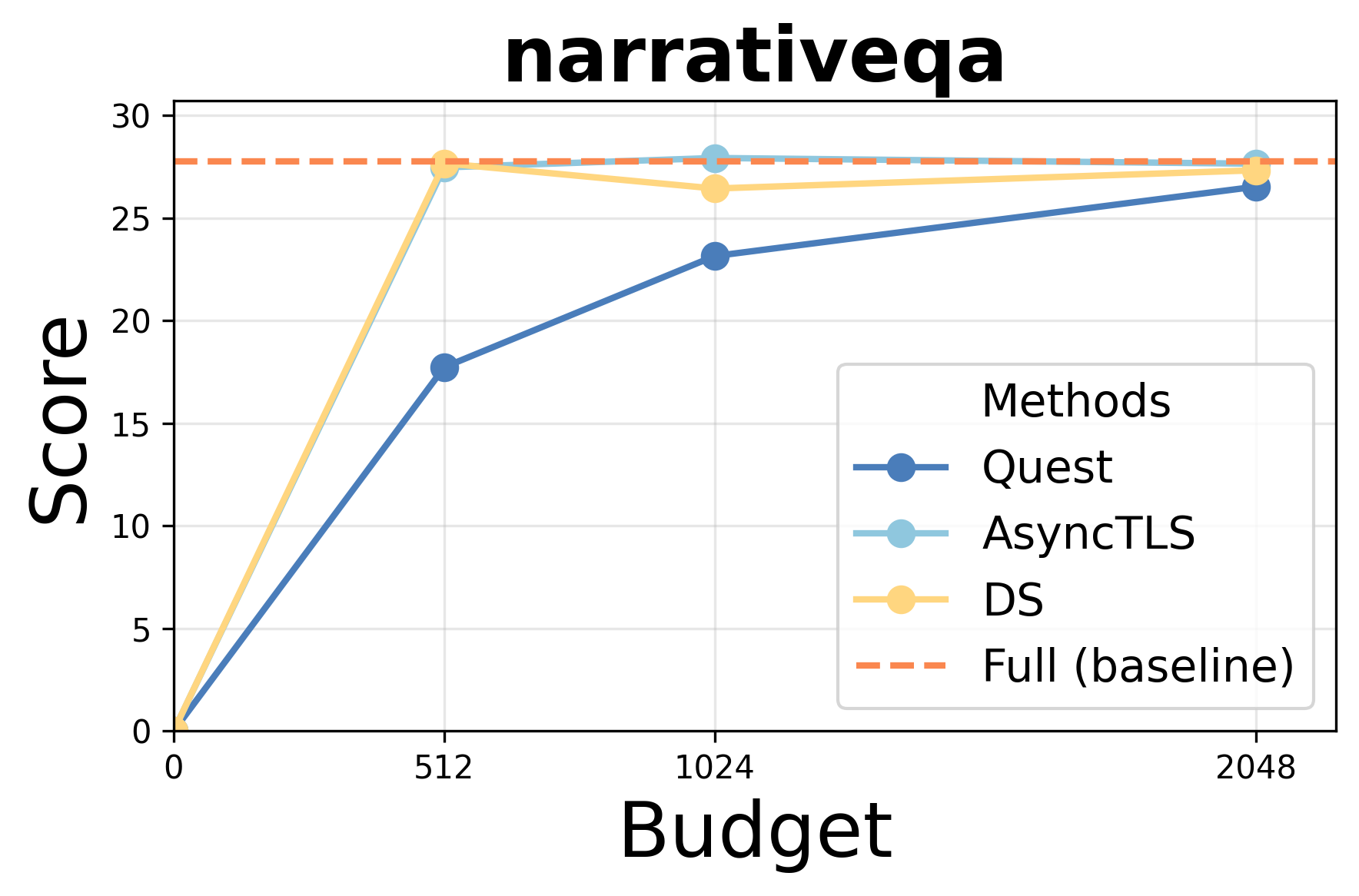}
    \end{minipage}
    \begin{minipage}[b]{0.23\textwidth}
        \centering
        \includegraphics[width=\textwidth]{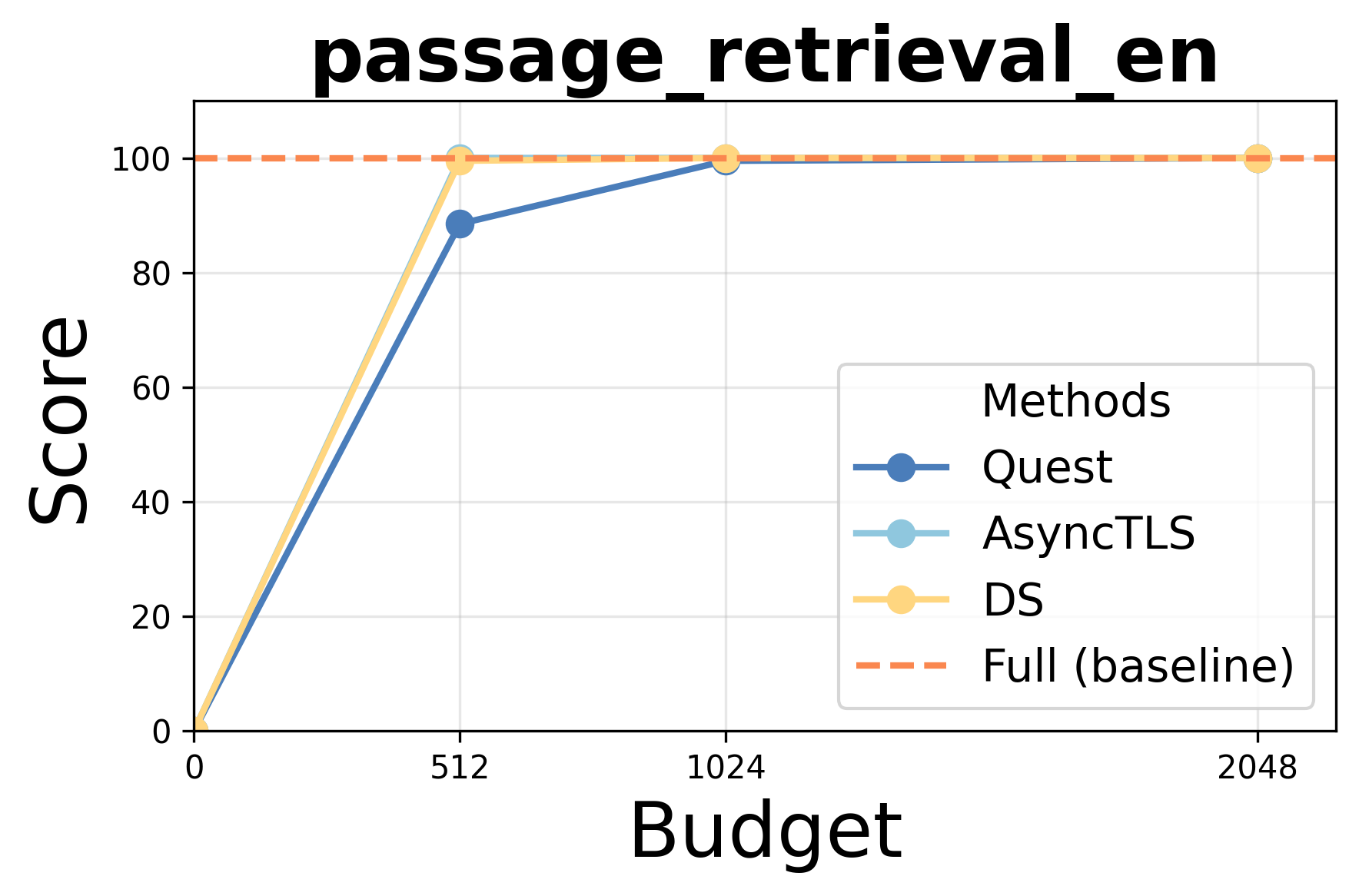}
    \end{minipage}
    \begin{minipage}[b]{0.23\textwidth}
        \centering
        \includegraphics[width=\textwidth]{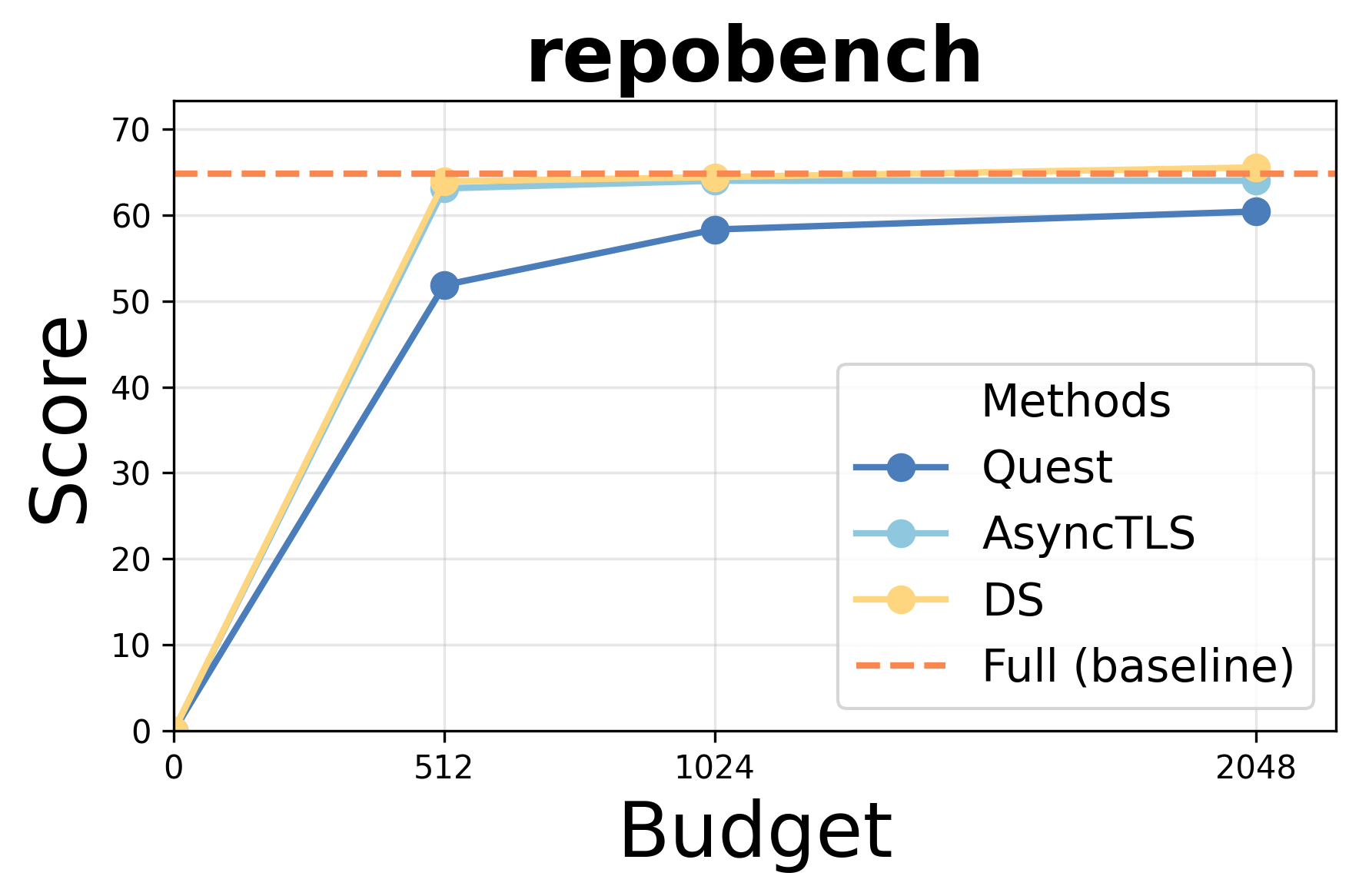}
    \end{minipage}
    \begin{minipage}[b]{0.23\textwidth}
        \centering
        \includegraphics[width=\textwidth]{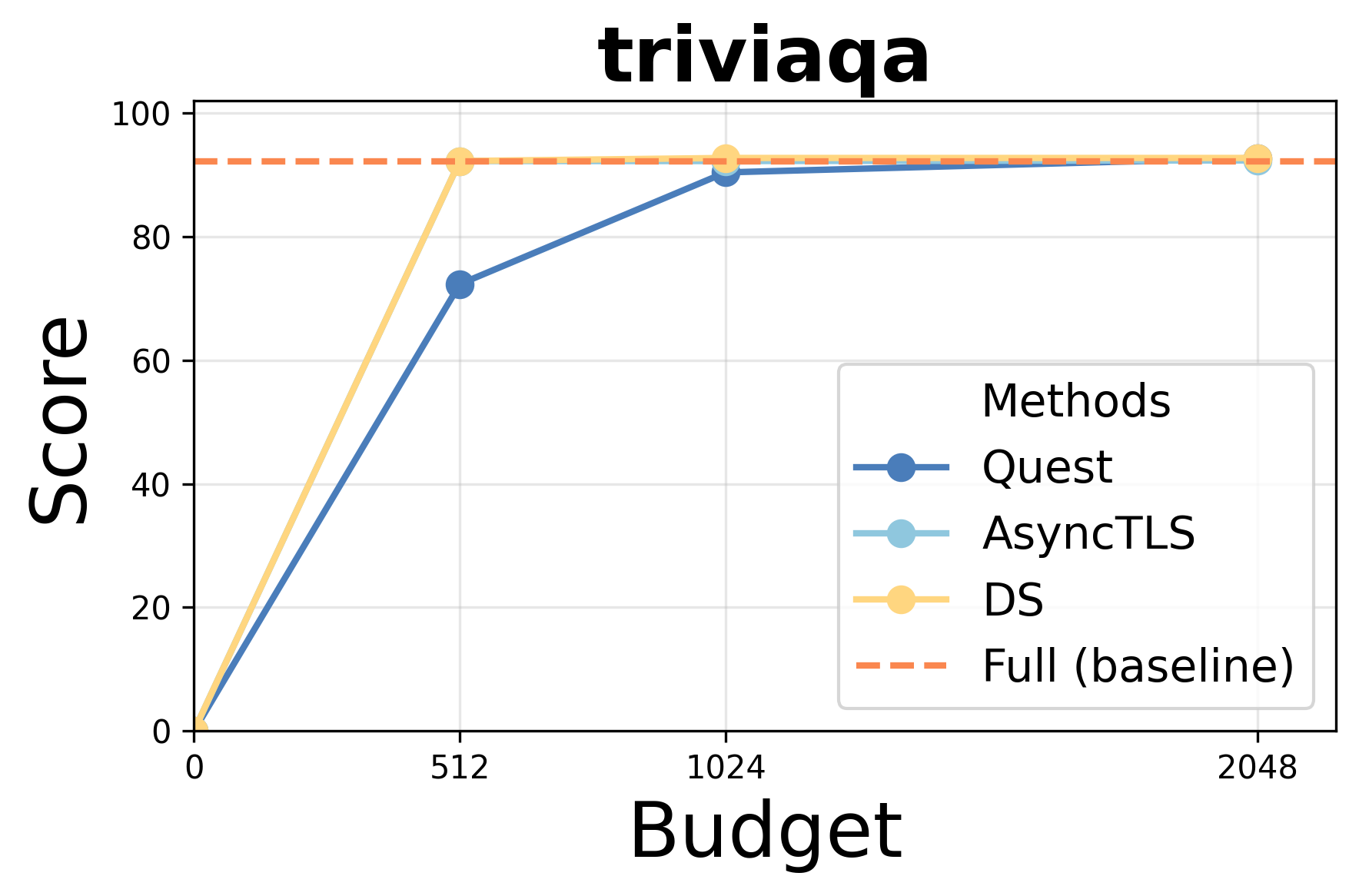}
    \end{minipage}
    
    \caption{Performance of AsyncTLS and baseline methods under various token budgets on Qwen3-14B.}
    \label{exp: token budgets}
\end{figure*}

\begin{figure*}[t]
    \centering
    \begin{minipage}[b]{0.23\textwidth}
        \centering
        \includegraphics[width=\textwidth]{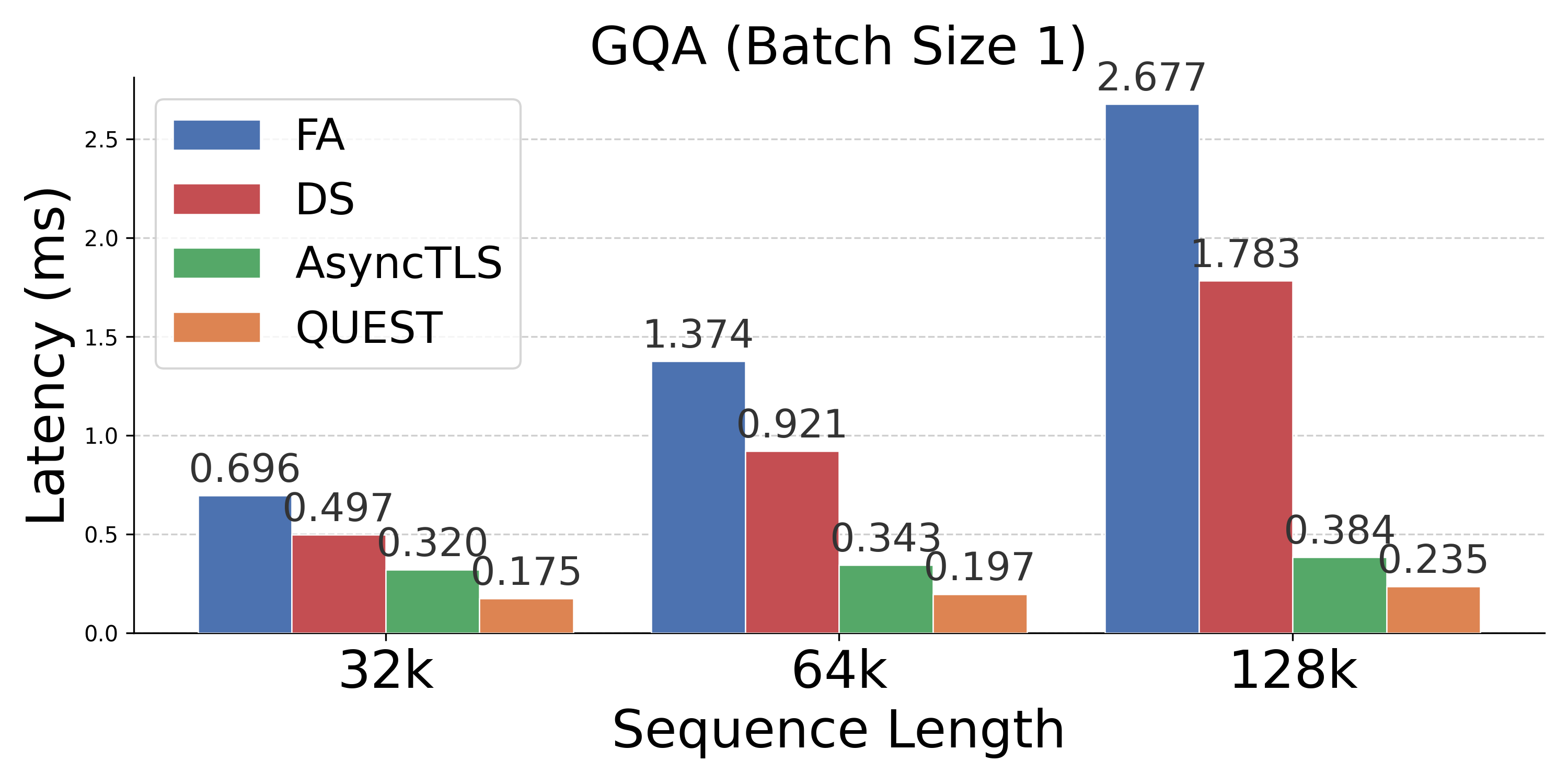}
    \end{minipage}
    \begin{minipage}[b]{0.23\textwidth}
        \centering
        \includegraphics[width=\textwidth]{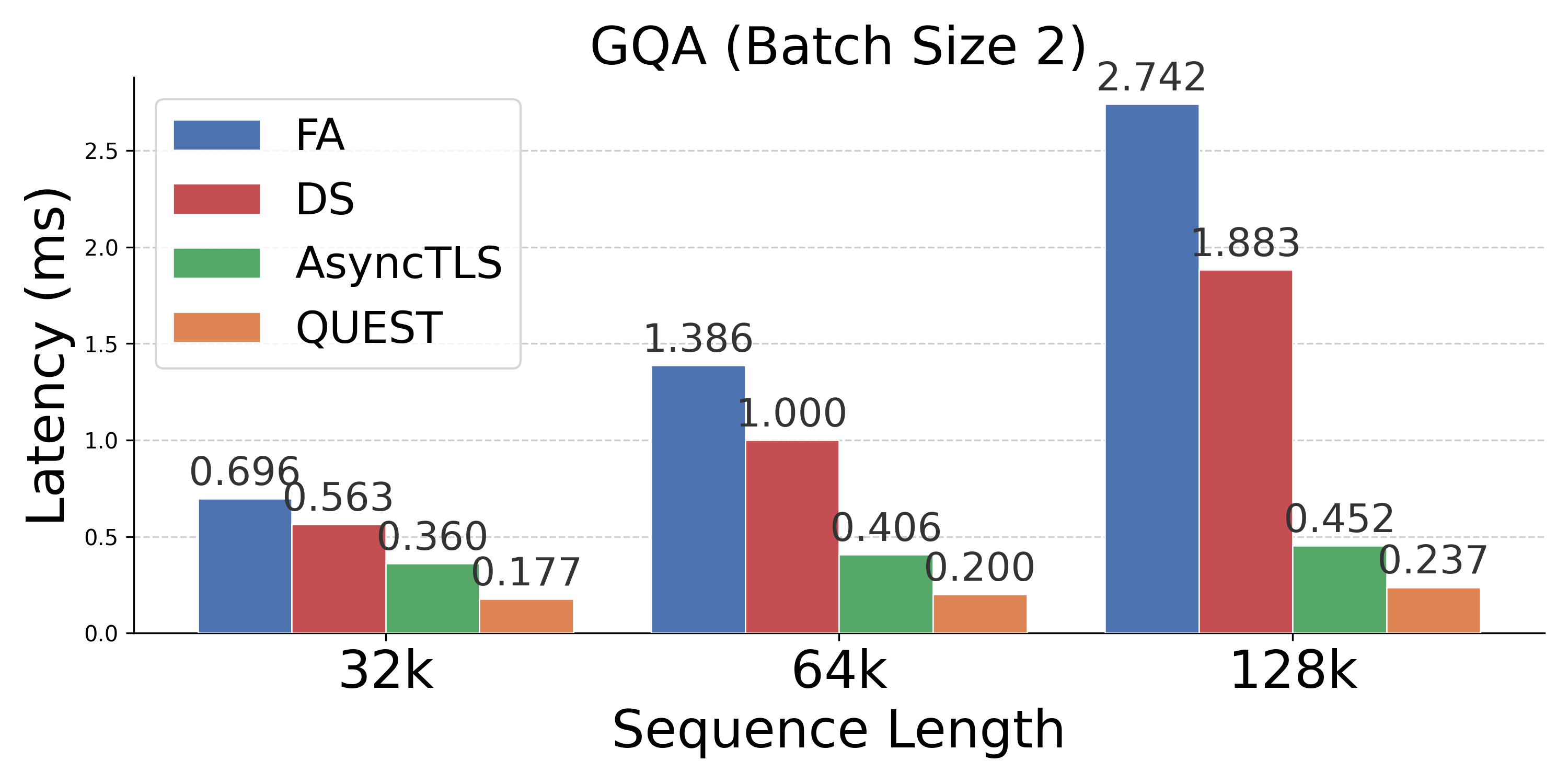}
    \end{minipage}
    \begin{minipage}[b]{0.23\textwidth}
        \centering
        \includegraphics[width=\textwidth]{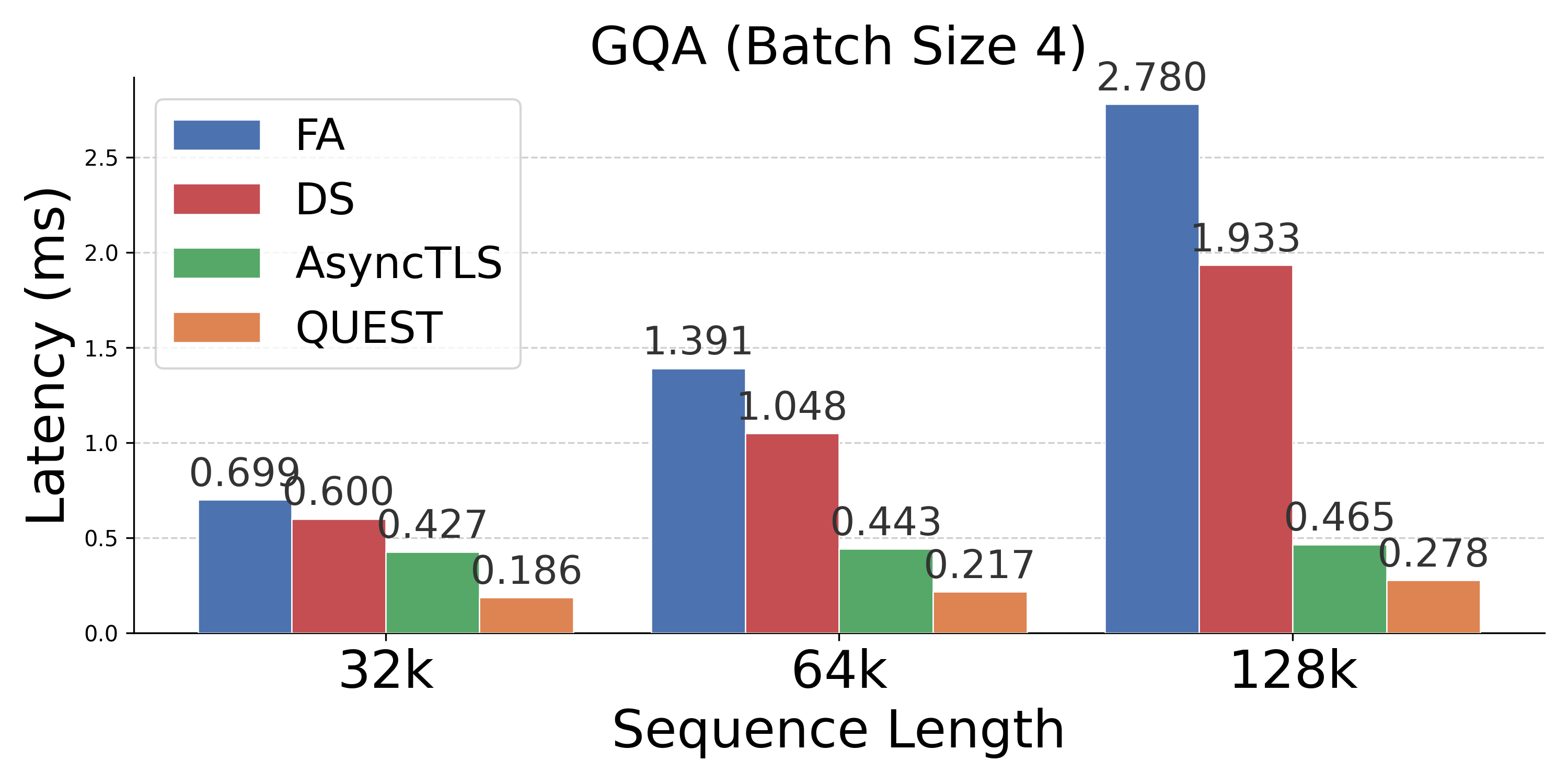}
    \end{minipage}
    \begin{minipage}[b]{0.23\textwidth}
        \centering
        \includegraphics[width=\textwidth]{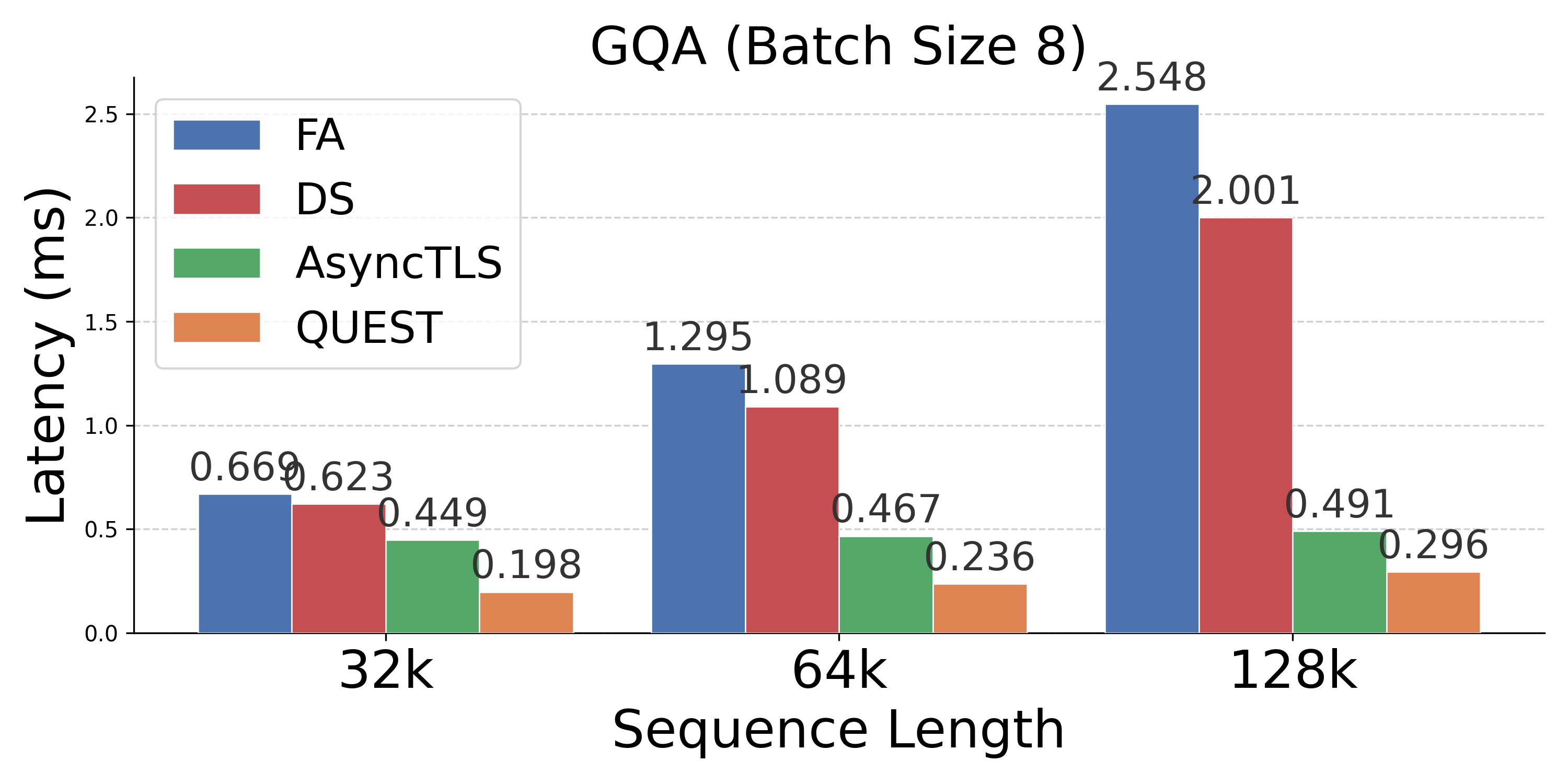}
    \end{minipage}
    \\
    \begin{minipage}[b]{0.23\textwidth}
        \centering
        \includegraphics[width=\textwidth]{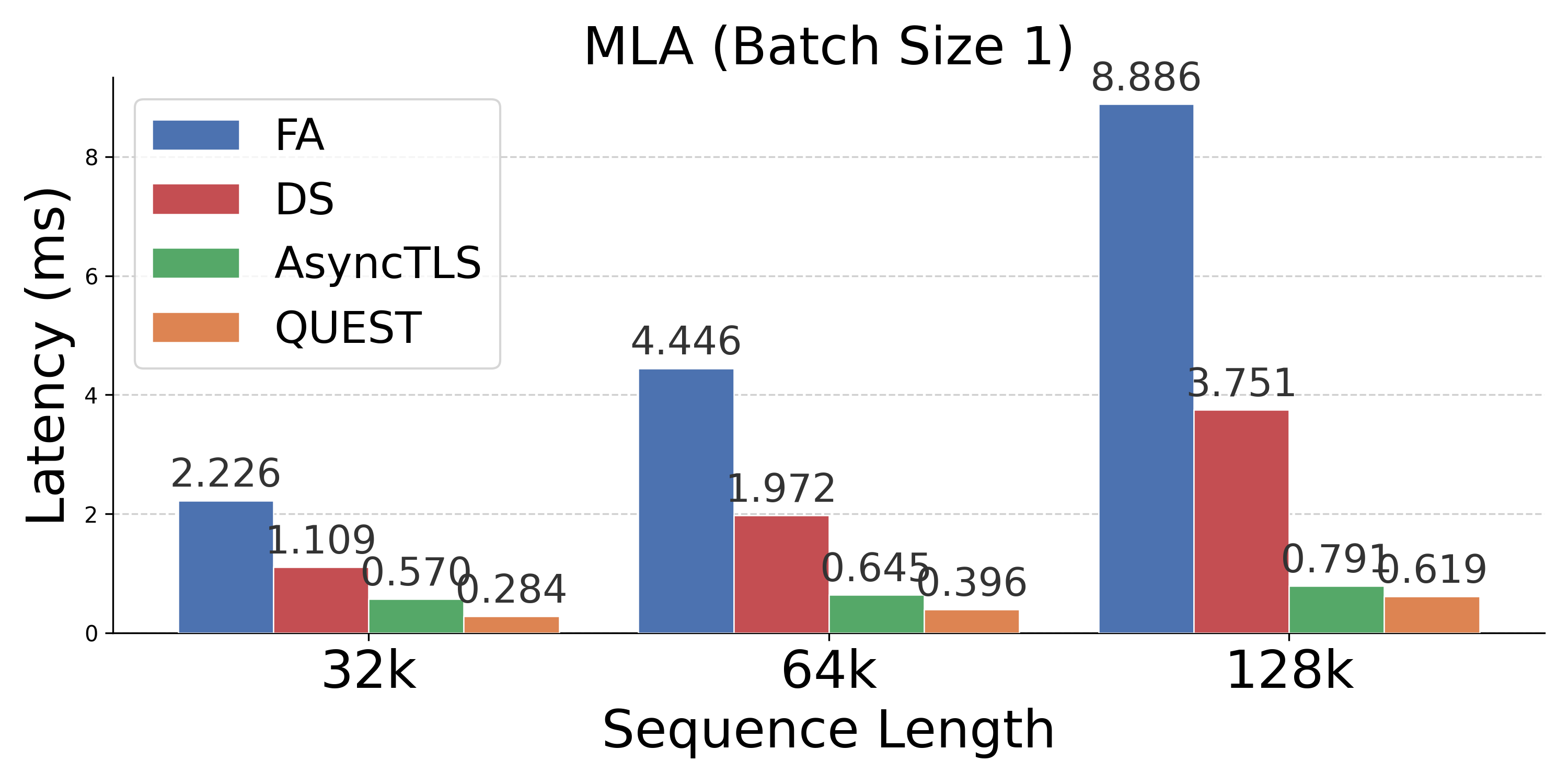}
    \end{minipage}
    \begin{minipage}[b]{0.23\textwidth}
        \centering
        \includegraphics[width=\textwidth]{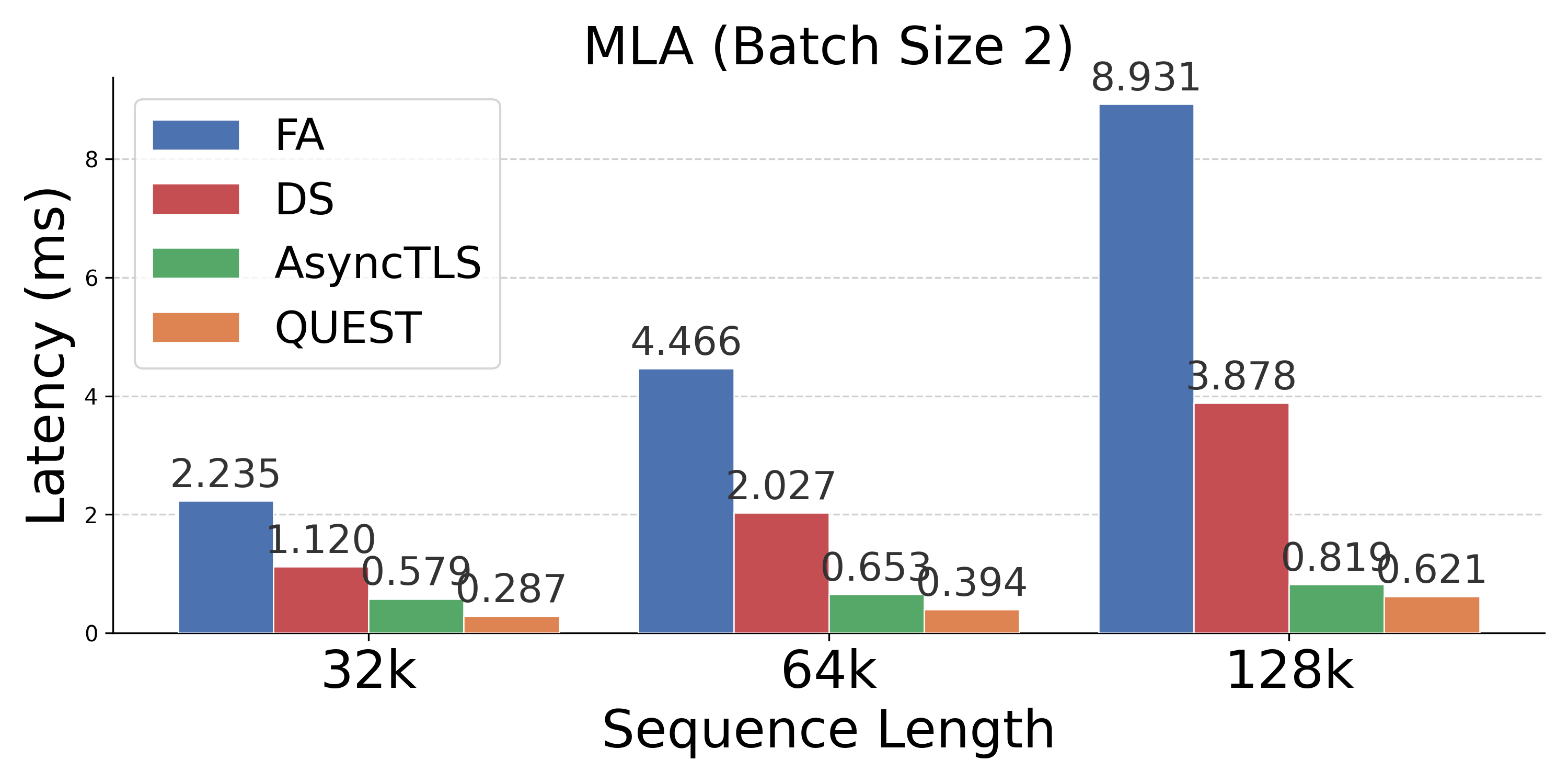}
    \end{minipage}
    \begin{minipage}[b]{0.23\textwidth}
        \centering
        \includegraphics[width=\textwidth]{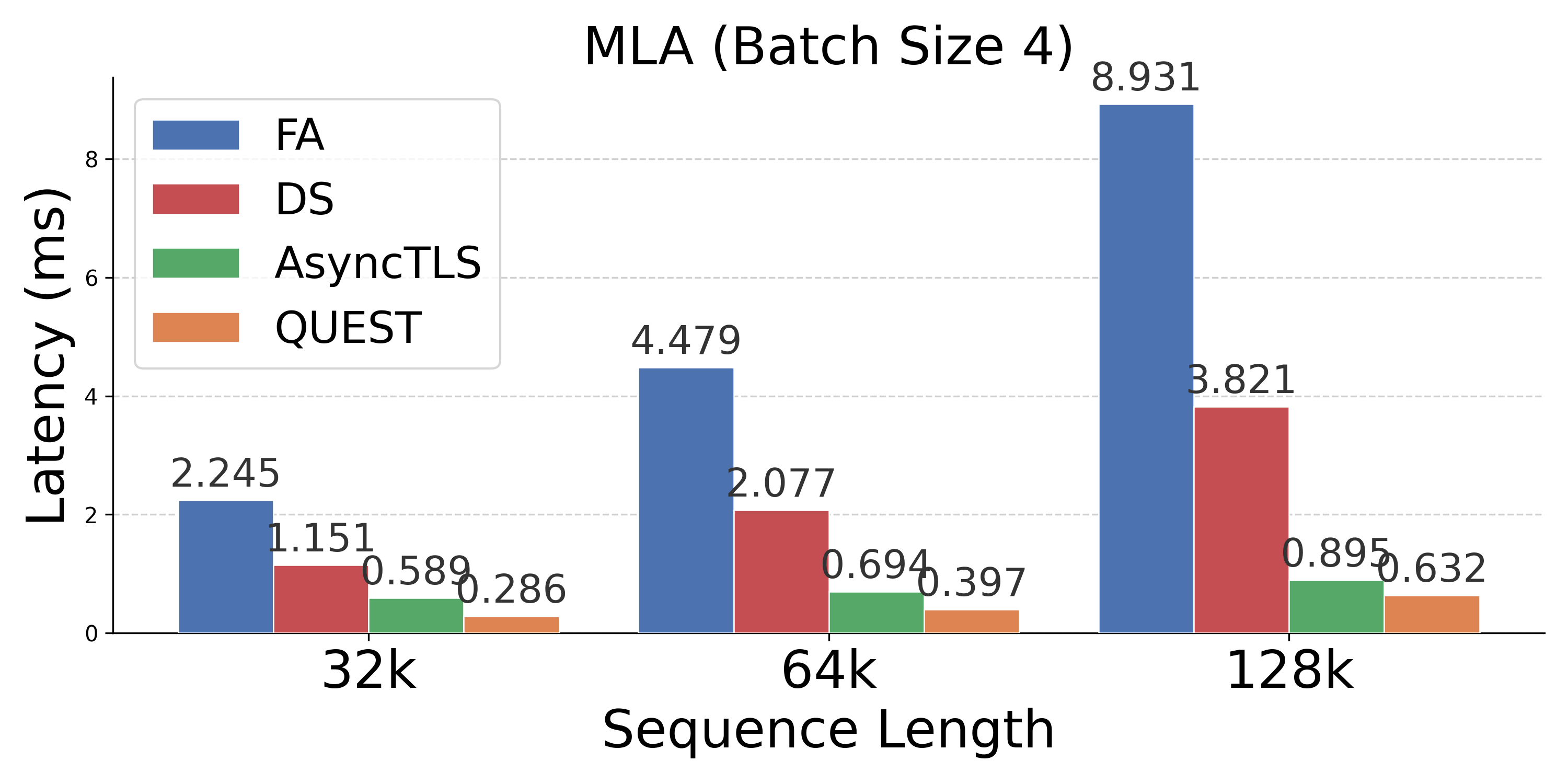}
    \end{minipage}
    \begin{minipage}[b]{0.23\textwidth}
        \centering
        \includegraphics[width=\textwidth]{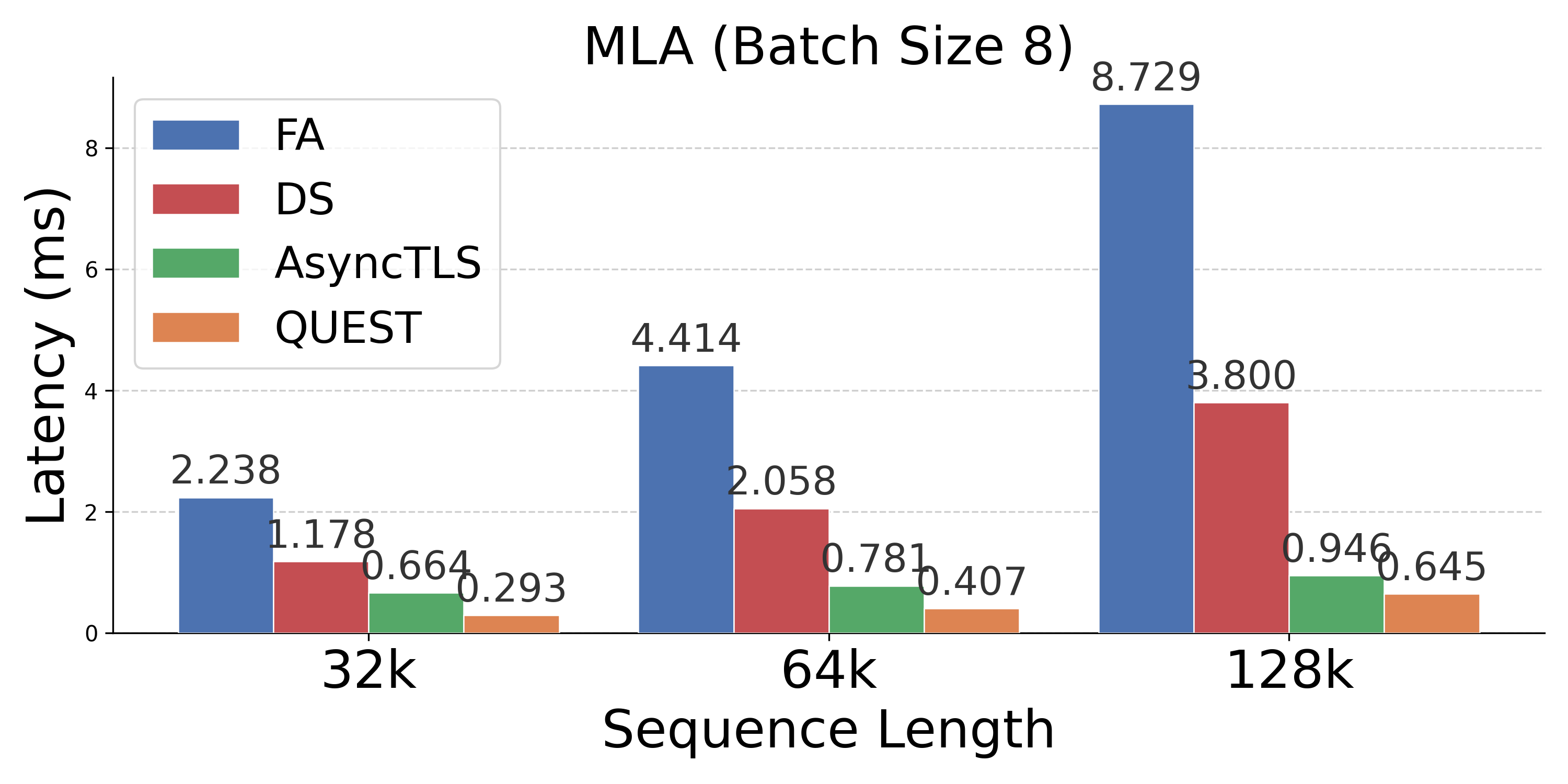}
    \end{minipage}
    \caption{Latency comparison of full attention (FA), token-level sparse attention (DS), block-level sparse attention (Quest), and two-level sparse attention (AsyncTLS) kernels across varying batch sizes and sequence lengths.\label{exp: latency}}
\end{figure*}

\begin{figure}[t]
    \centering
    \begin{minipage}[b]{\linewidth}
        \centering
        \includegraphics[width=\textwidth]{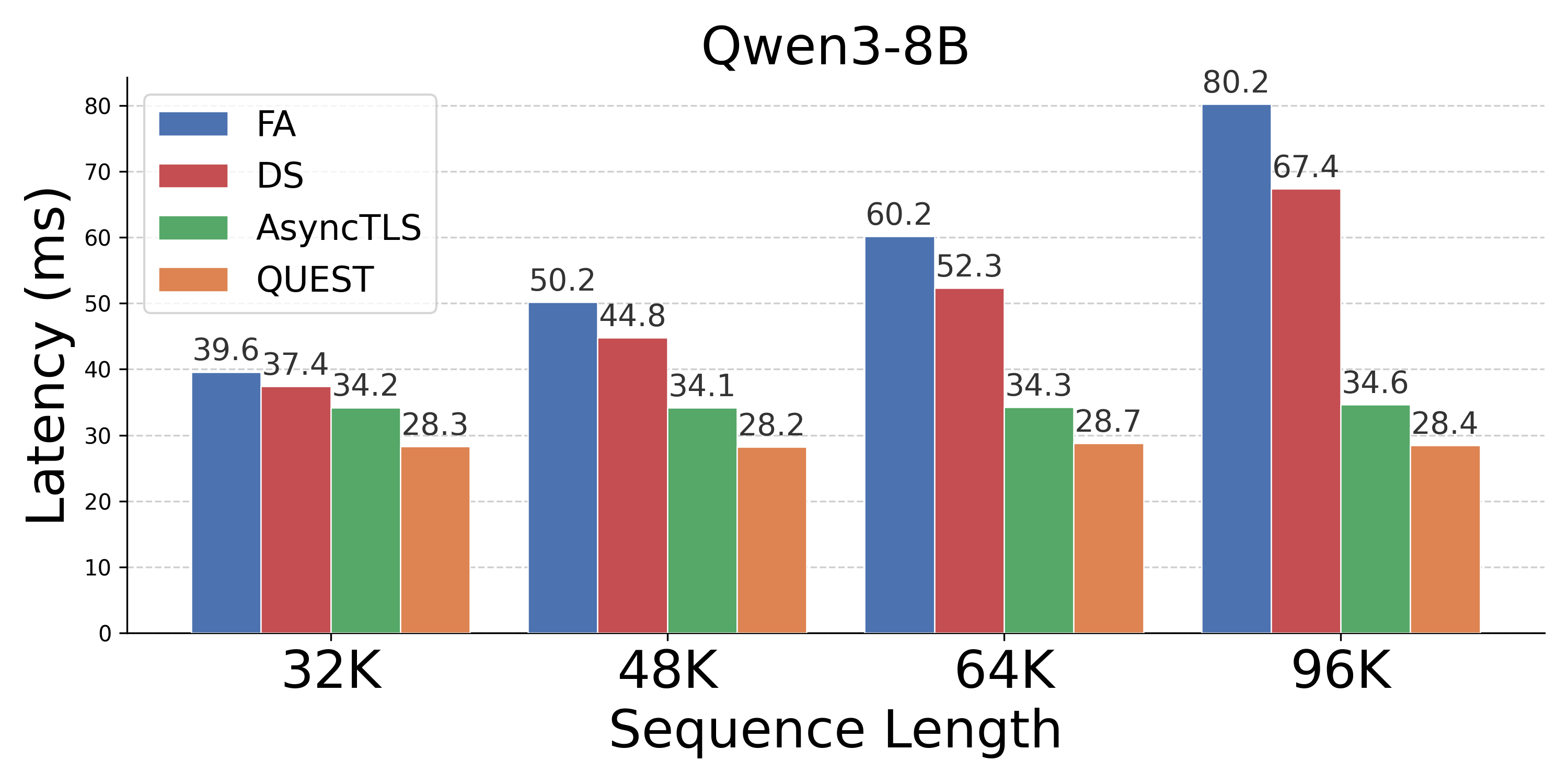}
    \end{minipage}
    \\
    \begin{minipage}[b]{\linewidth}
        \centering
        \includegraphics[width=\textwidth]{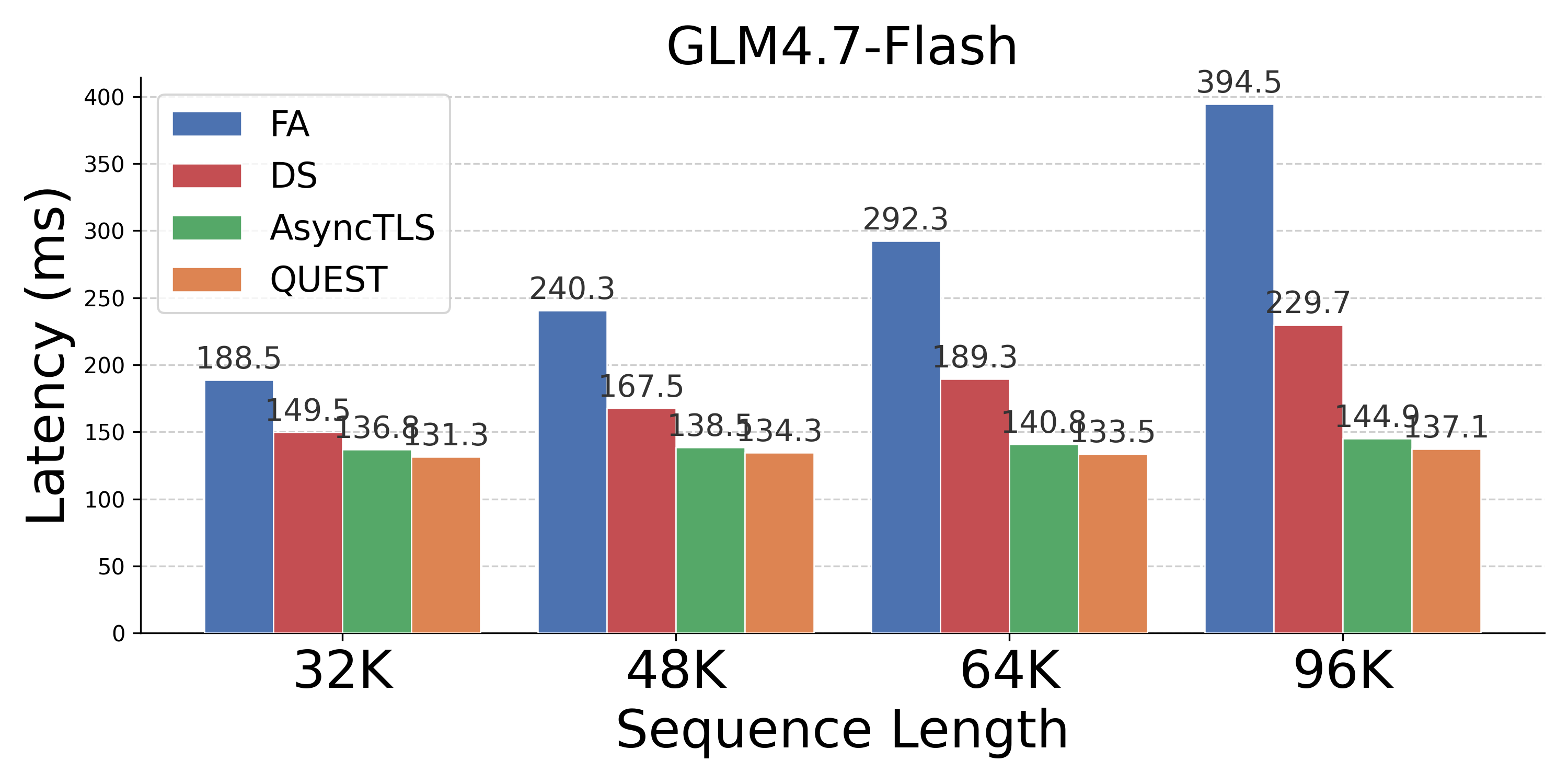}
    \end{minipage}

    \caption{End-to-end latency comparison of Qwen3-8B and GLM4.7-Flash with different attention methods across varying sequence lengths.\label{exp: latency}}
\end{figure}

\begin{figure}[t]
    \centering
    \begin{minipage}[b]{\linewidth}
        \centering
        \includegraphics[width=\textwidth]{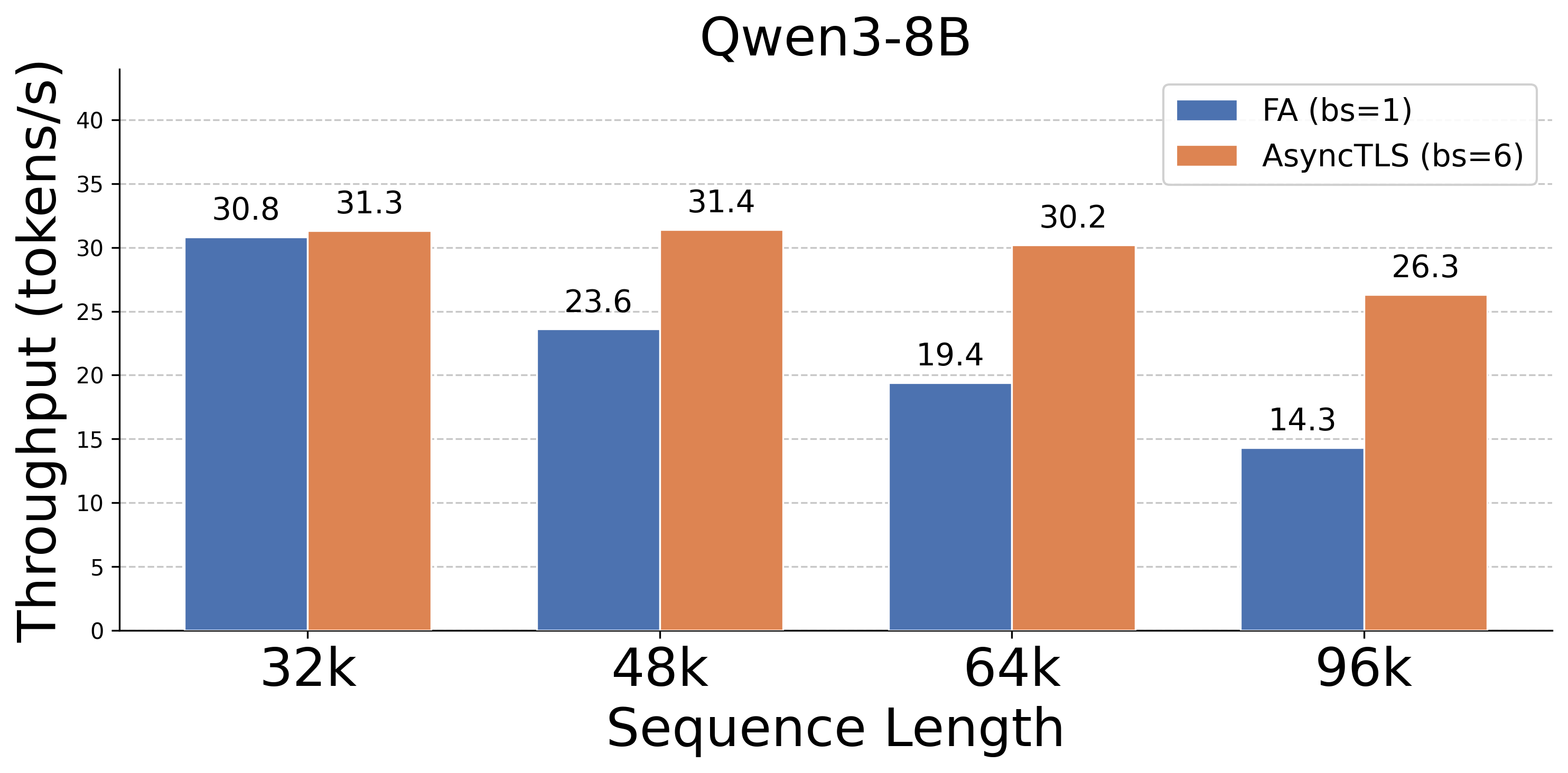}
    \end{minipage}
    \\
    \begin{minipage}[b]{\linewidth}
        \centering
        \includegraphics[width=\textwidth]{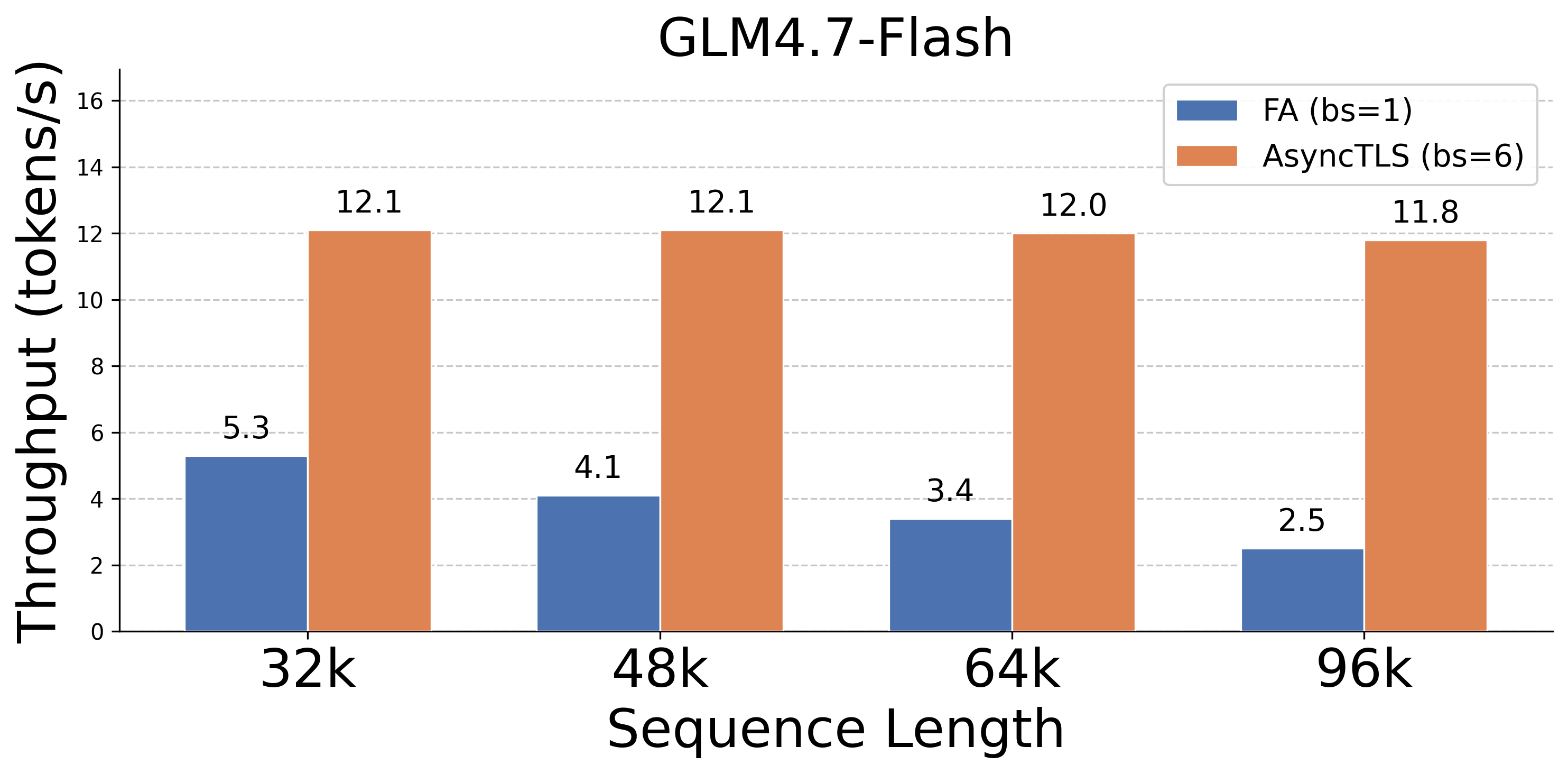}
    \end{minipage}
    
    \caption{Throughput comparison of Qwen3-8B and GLM4.7-Flash with full attention (FA) and AsyncTLS across varying sequence lengths.\label{exp: throughput}}
\end{figure}

\subsection{Hierarchical KV Cache Management}
\label{sec:async_offloading}

\paragraph{Prefetching with Temporal Overlap.}
Although sparse attention alleviates computational costs in long-context scenarios, it does not mitigate the prohibitive memory footprint of the KV cache. Consequently, recent studies have explored combining sparse attention with KV cache offloading to jointly optimize computation and storage. Meanwhile, constrained by the limited CPU-GPU bandwidth, recent efforts have centered on minimizing data transfer volumes and overlapping memory access with computation to hide latency. In \textbf{AsyncTLS}, to enable efficient token-level sparse attention under KV cache offloading, we exploit the temporal stability of attention patterns across consecutive decoding steps. The key observation is that block-level selections exhibit high locality: $\mathcal{M}_{t-1} \approx \mathcal{M}_{t}$.

At timestep $t$, while computing attention using the fine-grained set $\mathcal{S}_t$ derived from previous coarse selection $\mathcal{M}_{t-1}$, we simultaneously:
\begin{enumerate}
    \item Execute the coarse block selection to determine $\mathcal{M}_t$ for the next timestep;
    \item Initiate asynchronous prefetch of blocks in $\mathcal{M}_t$ that are not present in local GPU memory.
\end{enumerate}

This pipelining creates a one-step lag between coarse selection and fine-grained computation, formally:
\begin{align*}
    \mathcal{M}_t = \text{BlockSelect}(\mathbf{q}_t, \mathbf{K}, \mathbf{V}), \\    
    \mathcal{S}_t = \text{TokenSelect}(\mathbf{q}_t, \mathcal{M}_{t-1}), 
\end{align*}
\noindent where the coarse selection $\mathcal{M}_{t-1}$ from timestep $t-1$ guides the token-level pruning at $t$. The block selection $\mathcal{M}_t$ is computed in parallel with attention over $\mathcal{S}_t$ and following a feed-forward network.

\paragraph{Incremental Block Transmission.}
To minimize PCIe bandwidth consumption, we exploit the similarity between consecutive coarse selections. Instead of transferring entire blocks, we maintain a resident cache $\mathcal{C}_t$ on the GPU and transfer only the difference between $\mathcal{M}_t$ and $\mathcal{M}_{t-1}$. The incremental transfer set $\mathcal{T}_t$ contains only new blocks required for timestep $t$:
\begin{align*}
    \mathcal{T}_t & = \mathcal{M}_t \setminus \mathcal{C}_t,\quad \mathcal{C}_{t+1} = \mathcal{M}_t
\end{align*}

This strategy reduces bandwidth requirements from $O(k_b \cdot B \cdot d)$ to $O(| \mathcal{T}_t| \cdot B \cdot d)$ per step, where $| \mathcal{T}_t| \ll k_b$ due to temporal locality in attention patterns.

\subsection{Complexity Analysis}
\label{sec:complexity}

At each decoding step, AsyncTLS incurs three components: (i) $O(\frac{n}{B} d )$ for coarse-grained block scoring, (ii) $O(k_b B |\mathcal{C}|)$ for fine-grained token-level Top-$K$ selection, and (iii) $O(k_t d)$ for sparse attention computation. The total complexity is substantially lower than full attention ($O(n d)$) and token-level sparse attention ($O(n |\mathcal{C}| + k_t d)$ under a long-context scenario.

The KV cache transfer overhead is $O(|\Delta_t| B d)$, where $|\Delta_t|$ denotes the number of blocks with changed selection status between consecutive steps. Exploiting temporal locality ($|\Delta_t| \approx \epsilon k_b$ with $\epsilon \ll 1$), this is significantly reduced compared to block-level offloading baselines requiring $O(k_b B d)$ per step.

\section{Experiments}

We evaluate AsyncTLS following standard protocols in the literature, benchmarking it on in-context retrieval and long-context understanding tasks against both full attention and representative sparse attention baselines: block-level methods (e.g., Quest~\cite{quest-tang2024}) and token-level approaches (e.g., Double-Sparsity~\cite{double-sparsity-yang2024}).

\paragraph{Experiment Setup.} We conduct extensive experiments to validate AsyncTLS using Qwen3-8B, Qwen3-14B, and GLM-4.7-Flash. Following prior work on sparse attention, we configure the block-level index with a block size of 64 and retrieve 128 blocks (equivalent to 8,192 tokens) per query. For token-level indexing, we set the dimension to 32 for GQA models and 128 for MLA architectures, applying INT4 quantization to compress the index footprint. To investigate the impact of retrieval granularity on model performance, we vary the token budget across three settings: 512, 1024, and 2048 tokens.

\paragraph{In-Context Retrieval.}  
For in-context retrieval tasks, we employ a subset from RULER~\cite{ruler-hsieh2024} benchmark, which comprises 10 tasks: niah-single-1 (S1), niah-single-2 (S2), niah-multikey-1 (MK1), niah-multikey-2 (MK2), niah-multiquery (MQ), niah-multivalue (MV), RULER-QA-Hoptpot (QA1), RULER-QA-SQuAD (QA1), RULER-VT (VT), RULER-FWE (FWE).

\paragraph{Long Context Understanding.}  
To evaluate long context understanding, we use 14 tasks from the LongBench~\cite{longbench-bai2024}. These tasks cover various aspects, including narrative comprehension~\cite{narrativeqa-kočiský2017} (Narrative QA), scientific understanding~\cite{qasperqa-dasigi2021} (QasperQA), multi-hop reasoning (MultiField QA, Hotpot QA~\cite{hotpotqa-yang2018}, 2WikiMulti QA~\cite{2wikimultiqa-ho2020}, Musique~\cite{musique-trivedi2022}), document summarization (GovReport~\cite{govreport-huang2021}, QMSum~\cite{qmsum-zhong2021}, MultiNews~\cite{multinews-fabbri2019}), as well as specialized tasks such as TRec~\cite{TREC-Li2002LearningQC}, Trivia QA~\cite{triviaqa-joshi2017}, SAMSum~\cite{samsum-gliwa-etal-2019}, LCC~\cite{LCC-mohler-etal-2016}, and RepoBench-P~\cite{repobench-liu2023}.

\paragraph{Experiment Results.}
We present the evaluation results of AsyncTLS and baseline methods on LongBench and RULER in Table~\ref{exp: longbench},~\ref{exp: ruler qwen}, and~\ref{exp: ruler glm}\footnote{We excluded S1, S2, MK1, and MK2 from the RULER evaluation for GLM models due to abnormal performance metrics of GLM-4.7-flash under the default lm-eval-harness~\cite{eval-harness} configuration.}, respectively. For RULER, the token budget for sparse attention is set to 512, and the context length is set to 32k, while for LongBench, the token budget is set to 1024. It can be seen that, under identical token budget constraints, AsyncTLS consistently achieves superior performance compared to block-level sparse attention methods such as Quest, while maintaining results on par with the Full Attention baseline.

To investigate the impact of varying token budgets on sparse attention efficacy, we conduct experiments on a subset of LongBench tasks with token budgets configured at 512, 1024, and 2048. The results are presented in Figure~\ref{exp: token budgets}, which demonstrate that AsyncTLS outperforms Quest while achieving comparable performance to DS under the same token budget.

\paragraph{Efficiency.}  To assess the efficiency of AsyncTLS, we evaluate both operator-level performance and end-to-end inference latency and throughput.

We conduct kernel-level benchmarking for attention mechanisms, configuring GQA and MLA with 32 attention heads (with a group size of 4 for GQA). Using TileLang~\cite{tilelang-wang2026}, we implement four attention variants: two-level sparse attention (AsyncTLS), token-level sparse attention (DS), block-level sparse attention (QUEST) and full attention (FA). We evaluate inference latency across batch sizes of 1–8 and context lengths of 32K–128K. As illustrated in Figure~\ref{exp: latency}, TLS delivers substantial speedups: \(1.7\times\)-\(6.2\times\) over FA and \(1.2\times\)-\(4.0\times\) over DS for GQA, while achieving \(3.3\times\)-\(10.0\times\) and \(1.9\times\)-\(4.0\times\) improvements over FA and DS, respectively, for MLA. Meanwhile, compared to QUEST, AsyncTLS also achieves 54\% and 68\% of its inference speed for GQA and MLA, respectively.

To evaluate end-to-end latency and throughput, we conducted benchmark tests on Qwen3-8B and GLM4.7-Flash across sequence lengths ranging from 32K to 96K. First, we measured the end-to-end latency of the models without enabling cache offloading. As shown in Figure~\ref{exp: latency}, AsyncTLS achieves superior inference speed compared to DS and approaches that of QUEST, delivering 2.3$\times$ and 2.7$\times$ improvements over FA on Qwen3-8B and GLM4.7-Flash, respectively. Second, we evaluated the end-to-end throughput with cache offloading enabled. Leveraging the reduced KV cache footprint achieved through offloading techniques, AsyncTLS supports larger batch processing (batch size of 6), whereas the full attention mechanism is constrained to a batch size of 1. As illustrated in Figure~\ref{exp: throughput}, at a sequence length of 96K, AsyncTLS achieves 1.84$\times$ and 4.70$\times$ higher throughput than FA on Qwen3-8B and GLM4.7-Flash, respectively, highlighting its efficiency advantages in long-context scenarios.


\section{Conclusion}

Long-context LLM inference faces a fundamental tension between the accuracy of fine-grained token selection and the efficiency of coarse-grained block processing. We present AsyncTLS, a hierarchical sparse attention system that bridges this gap through a two-level architecture combining coarse block filtering with precise token-level attention. By staggering block selection and token computation across timesteps, our asynchronous offloading engine exploits temporal locality to overlap KV cache transfers with computation, minimizing PCIe bandwidth bottlenecks. 

Extensive evaluation across GQA and MLA architectures demonstrates that AsyncTLS achieves accuracy comparable to full attention while delivering substantial efficiency gains: $1.2\times$--$10.0\times$ operator speedups and $1.3\times$--$4.7\times$ end-to-end throughput improvements on contexts up to 96k tokens. These results establish that training-free token-level sparsity can be practically deployed with hardware-efficient indexing and hierarchical memory management, offering a scalable solution for ultra-long sequence generation.


% ...

\bibliography{custom}





\end{document}